\documentclass{article}

\usepackage[preprint]{neurips_2025}

\usepackage{booktabs}
\usepackage{amssymb}
\usepackage{pifont} 
\PassOptionsToPackage{numbers, compress}{natbib}
\usepackage[utf8]{inputenc} 
\usepackage[T1]{fontenc}    
\usepackage{hyperref}       
\usepackage{url}            
\usepackage{booktabs}       
\usepackage{amsfonts}       
\usepackage{nicefrac}       
\usepackage{microtype}      
\usepackage[dvipsnames,table]{xcolor}         
\usepackage{graphicx}
\usepackage{array}
\usepackage{comment}
\usepackage{amsmath}
\usepackage{tabularx}
\usepackage{tikz}
\usepackage{siunitx} 
\usepackage{caption} 
\usepackage{adjustbox} 
\usepackage{subcaption}
\usepackage{wrapfig}

\usepackage{multirow, makecell} 

\hypersetup{
    colorlinks=true,
    linkcolor=blue,
    filecolor=blue,      
    urlcolor= black,
    citecolor=MidnightBlue,
    pdftitle={Overleaf Example},
    pdfpagemode=FullScreen,
    }

\renewcommand{\arraystretch}{1.3}
\newcommand{\phaseicon}[1]{%
  \tikz[baseline=0.5ex] \draw[fill=#1, draw=black!30] (0,0) rectangle (0.3,0.3);%
}
\definecolor{pregrasp}{RGB}{255,225,235}     
\definecolor{contact}{RGB}{255,245,220}      
\definecolor{grasp}{RGB}{230,250,230}        
\definecolor{release}{RGB}{220,235,255}      
\definecolor{postgrasp}{RGB}{235,220,245}     

\newtheorem{Def}{Definition}[section]

\usepackage{xspace}

\newcommand{\benchname}{Robo2VLM\xspace}
\newcommand{\dsname}{Robo2VLM-1\xspace}

\usepackage[font=footnotesize]{caption}

\usepackage{xcolor,colortbl}          
\definecolor{spatclr}{RGB}{215,226,244}   
\definecolor{goalclr}{RGB}{208,235,200}   
\definecolor{intclr} {RGB}{230,222,247}   

\usepackage{tabularx}
\usepackage{pifont} 
\usepackage[most]{tcolorbox}
\tcbset{colback=gray!10, colframe=gray!80!black, boxrule=0.5pt, arc=1mm, left=1mm, right=1mm, top=1mm, bottom=1mm, boxsep=1mm}

\usepackage{float}
\usepackage{enumitem}

\title{Robo2VLM:\\ Visual Question Answering from Large-Scale In-the-Wild  Robot Manipulation Datasets}

\usepackage{hyperref}
\hypersetup{
    colorlinks=true,
    urlcolor=WildStrawberry
}

\author{
\textbf{Kaiyuan Chen}$^{1,*}$ ~~
\textbf{Shuangyu Xie}$^{1,*}$ ~~
\textbf{Zehan Ma}$^1$ ~~
\textbf{Pannag R Sanketi}$^2$ ~~
\textbf{Ken Goldberg}$^1$ \\
$^1$University of California, Berkeley \quad
$^*$Equal contribution \\
\texttt{\{kych, 
syxie, 
zehanma, 
goldberg\}@berkeley.edu} \\
\url{https://berkeleyautomation.github.io/robo2vlm/}
}

\begin{document}

\maketitle

\begin{abstract}
Vision-Language Models (VLMs) acquire real-world knowledge and general reasoning ability through Internet-scale image-text corpora. They can augment robotic systems with scene understanding and task planning, and assist visuomotor policies that are trained on robot trajectory data. We explore the reverse paradigm — using rich, real, multi-modal robot trajectory data to enhance and evaluate VLMs. In this paper, we present Robo2VLM, a Visual Question Answering (VQA) dataset generation framework for VLMs. Given a human tele-operated robot trajectory, Robo2VLM derives ground-truth from non-visual and non-descriptive sensory modalities, such as end-effector pose, gripper aperture, and force sensing. Based on these modalities, it segments the robot trajectory into a sequence of manipulation phases. At each phase, Robo2VLM uses scene and interaction understanding to identify 3D properties of the robot, task goal, and the target object. The properties are used to generate representative VQA queries – images with textural multiple-choice questions – based on spatial, goal-conditioned, and interaction reasoning question templates. We curate Robo2VLM-1, a large-scale in-the-wild dataset with 684,710 questions covering 463 distinct scenes and 3,396 robotic manipulation tasks from 176k real robot trajectories. Results suggest that Robo2VLM-1 can benchmark and improve VLM capabilities in spatial and interaction reasoning.
\end{abstract}

\section{Introduction}



Emerging Vision-Language Models (VLMs)~\cite{radford2021learning,qwen2.5,touvron2023Llama2,liu2023visual,anthropic2024claude35,openai2024gpt4o,google2025gemini25} can perform high-level reasoning and scene interpretation~\cite{Chen_2024_CVPR,karamcheti2024prismatic}. 
Recent robotic manipulation systems that integrate VLMs demonstrate enhanced capabilities in semantic and long horizon task reasoning \citep{kim24openvla,geminiroboticsteam2025geminiroboticsbringingai,shi2025hirobotopenendedinstruction}.
Yet, \textit{the} key challenge persists: the image-text corpora used for VLM pre-training high-quality lack fine-grained spatial information, which are prerequisites for robots to identify long-tail objects, complex scenes, reason about spatial relationships, and plan physical interactions.

To address this challenge, some research~\cite{islam2024eqamx, yang2025embodiedbench, li2024eai} relies on data generation through simulation~\cite{shridhar2020alfred, szot2021habitat, kolve2017ai2thor}.
However, such data has inherent limitations due to the sim-to-real gap, because simulator cannot accurately model visual properties such as noise, clutter, and lighting variations and physical properties such as contact dynamics, and interactions. Therefore, strong performance in simulation often fails to translate reliably to the physical world.
Meanwhile, deriving spatial knowledge from real-world (``in-the-wild'') data typically requires extensive and costly human labeling~\cite{sermanet2023robovqa,ji2025robobrain}.
In contrast, teleoperated robot trajectories that are used to train visuomotor policies~\cite{levine2016end}, such as Vision-Language-Action(VLA)~\cite{kim24openvla,octo_2023} or diffusion policies~\cite{chi2023diffusionpolicy}, typically include precise, structured proprioceptive and kinematic information—joint angles, end-effector poses, gripper states, and force–torque readings—that implicitly encode 3D spatial information. We hypothesize that visual and textual data extracted from robot trajectories can  improve VLM's spatial reasoning capabilities.

We present \benchname, a multiple-choice Visual Question Answering (VQA) dataset generation framework for VLMs from real-world robot data. Given a human-teleoperated robot trajectory, \benchname segments the trajectory into distinct manipulation phases, selects representative frames from each phase, and generates questions whose answers are supported by synchronized proprioceptive and kinematic ground truth.
We apply \benchname to 176k diverse, real-world trajectories from the Open X-Embodiment (OXE) dataset~\cite{open_x_embodiment_rt_x_2023}, producing over 3 million VQA samples. 
Inspired by data optimization paradigms such as domain reweighting in natural language processing~\cite{xie2023doremi} and robot policy learning~\cite{pmlr-v270-hejna25a}, we curate \dsname, a large-scale, in-the-wild VQA dataset with 684,710 questions covering 463 distinct scenes, 3,396 robotic manipulation tasks, and 149 manipulation skills.

%


We evaluate 14 model configurations with state-of-the-art open source models (LLaVA, Llama and Qwen) and with different parameter sizes and prompting techniques. The results indicate that some VLMs can achieve near human performance in questions related to object reachability and interaction understanding.  Evaluation also suggests a significant gap to human performance, especially in complex reasoning of fine-grained spatial relationship and interactions.
Finetuning LLaVA~\cite{liu2023visual} with \dsname improves most of the spatial and interaction capabilities with increasing training dataset size, with a maximum 50\% accuracy gain in state reasoning and task understanding. 

This paper makes the following contributions: (1) \benchname, a VQA data generation framework from real robot trajectories. (2) \dsname, an open VQA dataset with 684,710 questions covering diverse and realistic evaluation scenarios for manipulation. (3) Extensive evaluation data on state-of-the-art and fine-tuned VLMs.

\begin{figure}
    \centering
    \includegraphics[width=\linewidth]{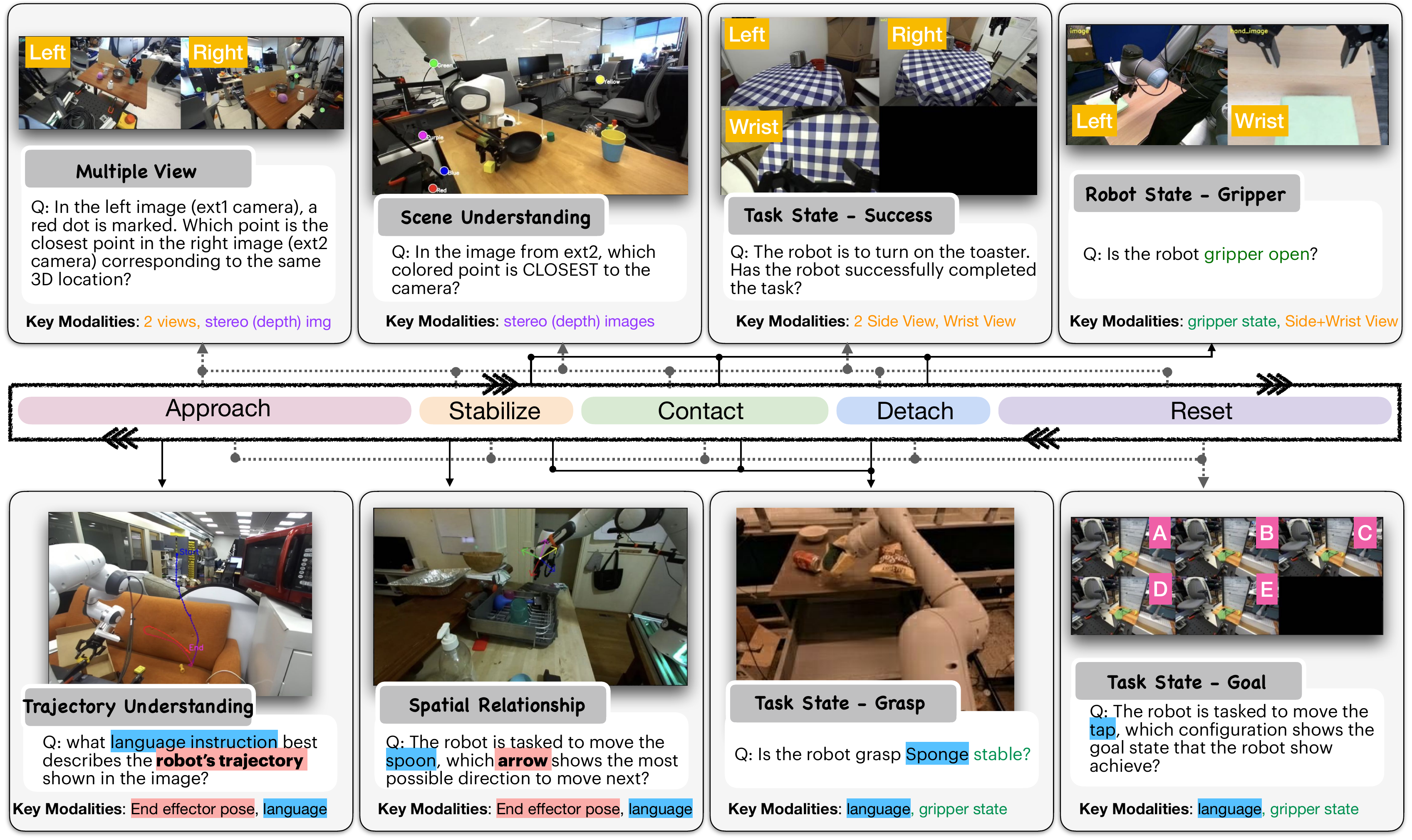}
    \caption{\textbf{\dsname dataset overview}. The middle colorbar traces a typical manipulation episode—from pre-grasp through immobilization, contact, detach, and into post-grasp. Surrounding panels give example questions for each VQA category. Dashed arrows connect every category to the phase(s) in which its questions are sampled. Icons beneath each panel list the key sensing modalities (RGB, stereo depth, wrist/side cameras, gripper state, end-effector pose, language instructions) needed to derive ground-truth answers. }
    \label{fig:vqa-examples}
    \vspace{-1em}
\end{figure}

\section{Related Work}

\paragraph{Large-Scale Robotics Datasets}
Recent large-scale robotics datasets, such as {Open‑X‑Embodiment}~\cite{open_x_embodiment_rt_x_2023} and DROID~\cite{droid}, provide extensive teleoperated demonstrations of complex manipulation skills. These datasets are foundational for training modern generalist robot policies—including Octo~\cite{octo_2023}, RT-1~\cite{brohan2023rt1roboticstransformerrealworld}, RT-2~\cite{brohan2023rt2visionlanguageactionmodelstransfer}, OpenVLA~\cite{kim24openvla}, Gemini Robotics~\cite{geminiroboticsteam2025geminiroboticsbringingai}, $\pi_0$~\cite{black2024pi0}, and Hi Robot~\cite{shi2025hirobotopenendedinstruction}—enabling them to learn diverse skills and understand nuanced physical interactions from broad data. Crucially for grounding VLMs, robotics datasets from {Open‑X‑Embodiment} contains rich sensory-modal including RGB video, proprioceptive
~\cite{khazatsky2024droid,brohan2023rt1roboticstransformerrealworld,rosete2022tacorl,mees23hulc2,lee2019icra,BerkeleyUR5Website,liu2022robot,Radosavovic2022,pari2021surprising,fanuc_manipulation2023,zhou2023modularity,zhou2023learning,zhu2022viola,zhu2022bottom,haldar2023watch},  
 depth data~\cite{khazatsky2024droid,brohan2023rt1roboticstransformerrealworld,rosete2022tacorl, mees23hulc2,BerkeleyUR5Website}, and force-torque~\cite{Radosavovic2022,fanuc_manipulation2023,zhou2023modularity, zhou2023learning}, that reflect the dynamics of interaction. These  information presents an opportunity to bridge robotics data with VLMs. 

\paragraph{VQA Benchmarks for Robotics and Embodied AI}
VQA offers a powerful paradigm for evaluating the visual reasoning capabilities of VLMs~\cite{antol2015vqa, goyal2017making, lee2022what}. Recently, VQA benchmarks have been developed for robotic tasks such as visual navigation in long-horizon planning \cite{das2018embodiedqa, anderson2018vision}. Simulation-based approaches~\cite{ islam2024eqamx, yang2025embodiedbench, li2024eai} (often utilizing environments like~\cite{shridhar2020alfred, szot2021habitat, kolve2017ai2thor}) generate large-scale VQA dataset, but face the persistent sim-to-real domain gap, where the result may not hold in reality due to factors like noise, clutter, and lighting variations. Real-world data benchmark, such as RoboVQA~\cite{sermanet2023robovqa} (human-verified Q/A), improve generalization to real world setting but often involve significant manual annotation effort. These methods typically do not fully automate VQA generation by exploiting the rich spectrum of non-visual modalities (e.g., force, torque, proprioception), limiting their ability to support questions grounded in concepts such as grasp stability or multi-view spatial alignment. In contrast, Robo2VLM reduces the need for manual annotation and enables interaction and physical properties reasoning that are underexplored in previous VQA benchmarks, such as gripper states, grasping stability, task goal, and spatial information focus on the robot and target objects.  

\vspace{0.5em}
\section{\benchname}


\benchname generates {five-way multiple-choice question answering (MCQ)} from real robot teleoperated trajectories. Robo2VLM offers the following key features:
(1) High-quality and representative keyframe selection from long-horizon, in-the-wild, multi-modal robot trajectories, ensuring semantic diversity and relevance;
(2) Manipulation-centric question generation encompassing spatial, goal-conditioned, and interaction reasoning, each aligned with specific manipulation phases and grounded in corresponding sensor modalities.

We begin by defining a robot trajectory as a time-synchronized sequence of data frames from multiple sensor modalities including exteroceptive and proprioceptive~\cite{DBLP:reference/robo/ChristensenH16}. Let $T$ denote the length of a trajectory, and let $t \in \{1, 2, \dots, T\}$ index the discrete time steps. 
\begin{Def}{(Robot Observation Data Frame)}
At each time step $t$, the robot data frame is represented as a tuple:
\[
\mathcal{D}_t = \left(\mathcal{I}_t^{\text{RGB}}, \mathcal{I}_t^{\text{Stereo}}, \mathbf{p}_t^{\text{EE}}, s_t^{\text{Gripper}}, \mathbf{f}_t \right)
\]
where $\mathcal{I}_t^{\text{RGB}} = \{I_t^{\text{RGB}} \in \mathbb{R}^{H \times W \times 3}\}$ is a set of multi-view RGB images captured from monocular cameras,
$\mathcal{I}_t^{\text{Stereo}} =\{ I_t^{\text{Stereo}} \in \mathbb{R}^{2 \times H \times W \times 3} \}$ denotes  a set of multi-view stereo image pair (left and right) if available, $\mathbf{p}_t^{\text{EE}} \in SE(3)$ is the 6-DoF end-effector pose and $s_t^{\text{Gripper}} \in \mathbb{R}$ denotes the scalar gripper state such as gripper aperture, $\mathbf{f}_t \in \mathbb{R}^6$ is the force-torque vector from the end-effector sensor. 
\end{Def}
The camera images are referred as exteroceptive sensing and the end-effector-related states belong to proprioceptive sensing.     

\begin{figure}
    \centering
    \includegraphics[width=\linewidth]{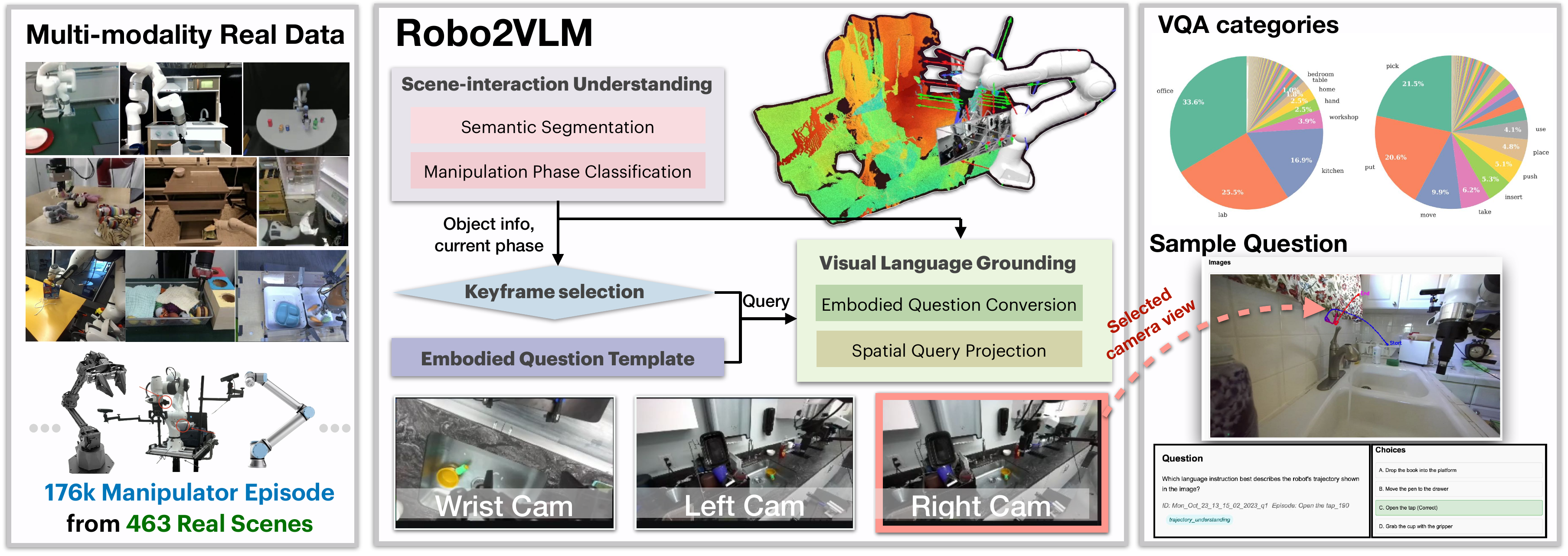}
    \caption{\textbf{\benchname framework.} Robo2VLM generates multi-modal real-world robot trajectories through (1) manipulation phase classification, (2) keyframe selection guided by scene and interaction cues, and (3) structured VQA question prototype.}
    \label{fig:benchmark_pipeline}
    \vspace{-1em}
\end{figure}

\begin{Def}{(Robot Trajectory)}
A trajectory $\mathcal{T}$ is defined as the temporally ordered sequence of observations $\mathcal{D}_{1:T}$ with a trajectory task language description $l$:
\[
\mathcal{T} = \left\{\mathcal{D}_{1:T} , l  \right\}
\]
\end{Def}
Given a robot trajectory, Robo2VLM (Fig.~\ref{fig:benchmark_pipeline}) begin with \textit{scene-interaction understanding}, applying semantic segmentation and manipulation phase classification to identify key segments (e.g., pre-grasp/approaching, contact, grasp, release). From these, we extract \textit{keyframes} based on phase transitions, scene coverage, and visibility of objects or the robot across multiple camera views. We use manipulation domain knowledge to design \textit{question prototype} to target core manipulation skills such as spatial relationship, goal conditions, and interaction understanding. \benchname instantiates these prototypes on selected keyframes and transforms them into natural language multiple-choice questions via a \textit{visual-language grounding} module that performs question conversion and spatial query projection.

\subsection{Scene-Interaction Understanding}

\paragraph{Embodied Scene Understanding}\label{sec:scene_understanding}
Given a task description in nature language and all images from different camera views, we first parse the language instruction using an off-the-shelf LLM such as Qwen 2.5 \cite{qwen2.5} to obtain \texttt{\small \{target object\}}, scene, task, and skill description.  
For the spatial understanding in manipulation, we need to know the relative direction and displacement between target object and gripper. From the proprioceptive data, we obtain the target object interaction point ground-truth from the robot trajectory data frames. 





\paragraph{Manipulation Phase Segmentation}\label{sec:manipulation}


To segment robotic manipulation trajectories into semantically meaningful phases, we define a temporal phase classification function based on the sequence of end-effector poses, gripper aperture signals, and force-torque measurements: $\mathbf{p}_{1:T}^{\text{EE}}, s_{1:T}^{\text{Gripper}}, \mathbf{f}_{1:T}$. To align different types of gripper aperture,  $s_t^{\text{Gripper}}$ is normalized to $[0, 1]$, where 0 indicates fully open and 1 indicates fully closed. Let \( s_t \in [0, 1] \) denote the normalized aperture at time \( t \), and \( \Delta s_t = s_t - s_{t-1} \) its temporal derivative. \( \Delta s_t \approx 0 \) denotes a small change within a tolerance margin \( \epsilon \), typically set to filter out noise. Let \( \|\mathbf{f}_t\| \) be the force magnitude (if available). 
We introduce three threshold parameters: \( \tau_g \) (grasp threshold), \( \tau_c \) (closure threshold), and \( \tau_f \) (force threshold for contact detection). 
Manipulation processes can be represented as a sequence of discrete phases, including approaching, stabilizing, contacting, releasing, and resetting or transitioning to subsequent actions. We denote the phase varible as
$
\Phi = \left\{ \Phi_{\text{app}}, \Phi_{\text{stab}}, \Phi_{\text{cont}}, \Phi_{\text{rel}}, \Phi_{\text{reset}}, \Phi_{\text{trans}} \right\}
$. Each timestep \( t \) is assigned a label \( \phi_t \in \Phi \) according to the following temporal logic rules:
\begin{align*}
\phi_t = 
\begin{cases}
\Phi_{\text{app}} & \text{if } s_t < \tau_g \land \Delta s_t < -\epsilon \\
\Phi_{\text{stab}} & \text{if } \phi_{t-1} = \Phi_{\text{app}} \land s_t < \tau_g \land |\Delta s_t| \leq \epsilon \\
\Phi_{\text{cont}} & \text{if } \phi_{t-1} = \Phi_{\text{stab}} \land s_t \geq \tau_c \land |\Delta s_t| \leq \epsilon \land (\|\mathbf{f}_t\| > \tau_f \lor \text{force unavailable}) \\
\Phi_{\text{rel}} & \text{if } \phi_{t-1} = \Phi_{\text{cont}} \land s_t \geq \tau_c \land \Delta s_t > \epsilon \\
\Phi_{\text{reset}} & \text{if } \phi_{t-1} = \Phi_{\text{rel}} \land s_t < \tau_g \land \Delta s_t > \epsilon \\
\Phi_{\text{trans}} & \text{otherwise}
\end{cases}
\end{align*}

The inclusion of force magnitude ensures that passive closure without external contact is not misclassified as active interaction. This multimodal phase labeling strategy captures both kinematic intent and physical contact, enabling robust segmentation of diverse manipulation behaviors.

To enforce a temporally coherent yet flexible phase progression, we define a partial order over the manipulation phases:
\[
\Phi_{\text{app}} \prec \Phi_{\text{stab}} \prec \Phi_{\text{cont}} \prec \Phi_{\text{rel}} \prec \Phi_{\text{reset}} \rightarrow \Phi_{\text{app}}
\]
This structure enforces unidirectional transitions along the phase chain, while allowing both phase skipping (e.g., directly from \(\Phi_{\text{app}}\) to \(\Phi_{\text{cont}}\)) and looping from the terminal phase \(\Phi_{\text{reset}}\) back to the initial phase \(\Phi_{\text{app}}\), which is common in sequential manipulation routines. At each time step \( t \), the phase label must satisfy \(\phi_t \succeq \phi_{t-1}\), or \(\phi_t = \Phi_{\text{app}}\) if \(\phi_{t-1} = \Phi_{\text{reset}}\), ensuring temporal monotonicity or task repetition without reversal. The auxiliary state \(\Phi_{\text{trans}}\) is used for ambiguous, missing, or conflicting observations where no confident assignment is possible. This symbolic, temporally-constrained model supports robust segmentation of complex manipulation behaviors under noisy or partially missing sensory input.

\begin{table}
  \centering
  \scriptsize
    \caption{Categorization of visual reasoning questions for robotic manipulation, with manipulation phase (color-coded) and data modality context. 
    \protect\phaseicon{pregrasp} Approach, 
    \protect\phaseicon{contact} Stabilize, 
    \protect\phaseicon{grasp} Contact, 
    \protect\phaseicon{release} Release, 
    \protect\phaseicon{postgrasp} Rest.}
  \setlength{\tabcolsep}{6pt}
  \renewcommand{\arraystretch}{1.1}

  \begin{tabular}{@{}%
      >{\raggedright\arraybackslash}p{2.5cm}p{6.2cm}  
      p{2.0cm}                                
      p{1.8cm}@{}                             
    }
    \toprule
    \textbf{Capabilities} & \textbf{Question Prototype} &
    \textbf{Manip. Phase} &
    \textbf{Sensor Modality} \\
    \midrule

    \rowcolor{gray!15}\multicolumn{4}{@{}c}{\textbf{Spatial Reasoning}} \\[.25em]

    Object State & Is the \texttt{\scriptsize\{target object\}} reachable by the robot? &
    ~\phaseicon{pregrasp} &
    $I_t^{\text{RGB}},\ D_t$ \\

    Spatial Relationship & What’s the relative direction in 3‑D between end effector and \texttt{\scriptsize\{target object\}}? &
    ~\phaseicon{pregrasp}~\phaseicon{contact} &
    $I_t^{\text{RGB}},\ \mathbf{p}_t^{\text{EE}}$ \\

    Scene Understanding & Which point is closer to the camera viewing the scene? &
    ~\phaseicon{pregrasp}~\phaseicon{contact} &
    $I_t^{\text{RGB}},\ I_t^{\text{Stereo}}$ \\

    Multiple View & Which point in the right‑side image corresponds to the point in the left‑side image? &
    ~\phaseicon{pregrasp}~\phaseicon{contact}~\phaseicon{release}~\phaseicon{postgrasp} &
    $I_t^{\text{Stereo}}$ \\

    \midrule
    \rowcolor{gray!15}\multicolumn{4}{@{}c}{\textbf{Goal-conditioned Reasoning}} \\[.25em]

    Task State-success & Has the robot successfully completed the task? &
    ~\phaseicon{postgrasp}~ &
    $I_t^{\text{RGB}}$ \\

    Task State-Goal & What is the goal configuration for \texttt{\scriptsize\{interaction\}}? &
    ~\phaseicon{pregrasp}~\phaseicon{contact}~\phaseicon{grasp}~\phaseicon{release}~&
    $I_t^{\text{RGB}},\ \mathbf{p}_t^{\text{EE}}$ \\

    Action Understanding & The robot is \texttt{\scriptsize\{interaction\}}. What is the robot's current action phase? & ~\phaseicon{pregrasp}~\phaseicon{contact}~\phaseicon{grasp}~\phaseicon{release}~\phaseicon{postgrasp}~ &
    $I_t^{\text{RGB}},\ \mathcal{T}_{1:t}$ \\

    Interaction Phase & What will the robot do next? &
    ~\phaseicon{pregrasp}~\phaseicon{contact}~\phaseicon{grasp}~\phaseicon{release}~&
    $I_t^{\text{RGB}},\ \dot{\mathbf{p}}_t^{\text{EE}}$ \\

   Trajectory Understanding &  What task does this trajectory likely accomplish? &
    ~\phaseicon{pregrasp}~ &
    $I_t^{\text{RGB}},\ \mathbf{p}_t^{\text{EE}}$ \\

    \rowcolor{gray!15}\multicolumn{4}{@{}c}{\textbf{Interaction Reasoning}} \\[.25em]

    Task State-grasp & Is this a stable grasp? &
    ~\phaseicon{contact}~\phaseicon{grasp}~\phaseicon{release}~ &
    $I_t^{\text{RGB}},\ \mathbf{f}_t$ \\

    Robot State & Is the robot gripper currently open? &
    ~\phaseicon{contact}~\phaseicon{grasp}~\phaseicon{release}~ &
    $I_t^{\text{RGB}},\ s_t^{\text{Gripper}}$ \\

    \bottomrule
  \end{tabular}
  \label{tab:reasoning-types}
  \vspace{-1em}
\end{table}

\subsection{Visual Question Prototype}

We design a set of \textit{visual question prototypes}, each of which aligns with specific manipulation task completion required robot capabilities and anchors to distinct manipulation phases as illustrated in Table~\ref{tab:reasoning-types}. 
These prototypes are organized into three reasoning categories. 

\textbf{Spatial Reasoning} focuses on the robot’s understanding of object geometry, reachability, and spatial layout across viewpoints. Questions such as ``Is the object reachable?'' or ``What’s the relative direction between the gripper and the object?'' are grounded in the early approach \phaseicon{pregrasp} and stabilize~\phaseicon{contact} stages. These rely on RGB, depth, stereo, and 3D gripper pose data, which together enable accurate localization and spatial inference across frames or views. 

\textbf{Goal-conditioned Reasoning} probes the agent’s high-level understanding of tasks, including goal inference, future action prediction, and overall task success. Questions such as ``Is the task failed?'', ``What will the robot do next?'', and ``What is the robot’s current action phase?'' span multiple manipulation phases from approach~\phaseicon{pregrasp} through reset~\phaseicon{postgrasp}. These require temporal context, pose estimation, and sometimes motion history, leveraging the multi-step evolution of the scene. 

\textbf{Interaction Reasoning} focuses on physical interaction dynamics, such as grasp stability or the robot’s current actuator state. These occur during stabilize~\phaseicon{contact}, contact~\phaseicon{grasp}, and release~\phaseicon{release} phases, and depend on RGB, tactile, or gripper aperture signals. For instance, the question ``Is this a stable grasp?'' may depend on contact force readings or inferred object displacement.

The ground truth of the questions are grounded by multiple sensor modality observations. We design the incorrect answers as part of the visual question prototypes. For example, in the scene understanding, we require the sampled points to be significantly different in depth from other points and from the depth sensor to account for sensor inaccuracy. In action understanding, the correct action arrow differs significantly from the distractor arrows by having a large angular separation in the projected 2D image.  To detect guessing by hallucination, we randomly replace some correct answers with "None of Above" option.




\subsection{Keyframe Selection}
\label{sec:keyframe_selection} 

Given that raw robotic trajectories often contain hundreds of frames sampled at high frequency, using all frames is computationally expensive and can introduce redundancy due to minimal temporal variation. Moreover, many intermediate frames are visually or semantically uninformative for downstream reasoning tasks. To address this, we select a compact set of keyframes that retain essential semantic and visual cues while reducing redundancy and data volume. These keyframes are extracted from the multi-modal robot trajectory $\mathcal{T} = \{\mathcal{O}_t\}_{t=1}^T$ based on manipulation phase transition, scene coverage diversity and context visibility.

    

\newcommand{\cmark}{\textcolor{green!60!black}{\checkmark}}
\newcommand{\xmark}{\textcolor{red!75!black}{\ding{55}}}
\newcommand{\unkmark}{\textcolor{gray}{---}}

\section{\dsname Dataset}

\begin{figure}
    \centering
    \includegraphics[width=\linewidth]{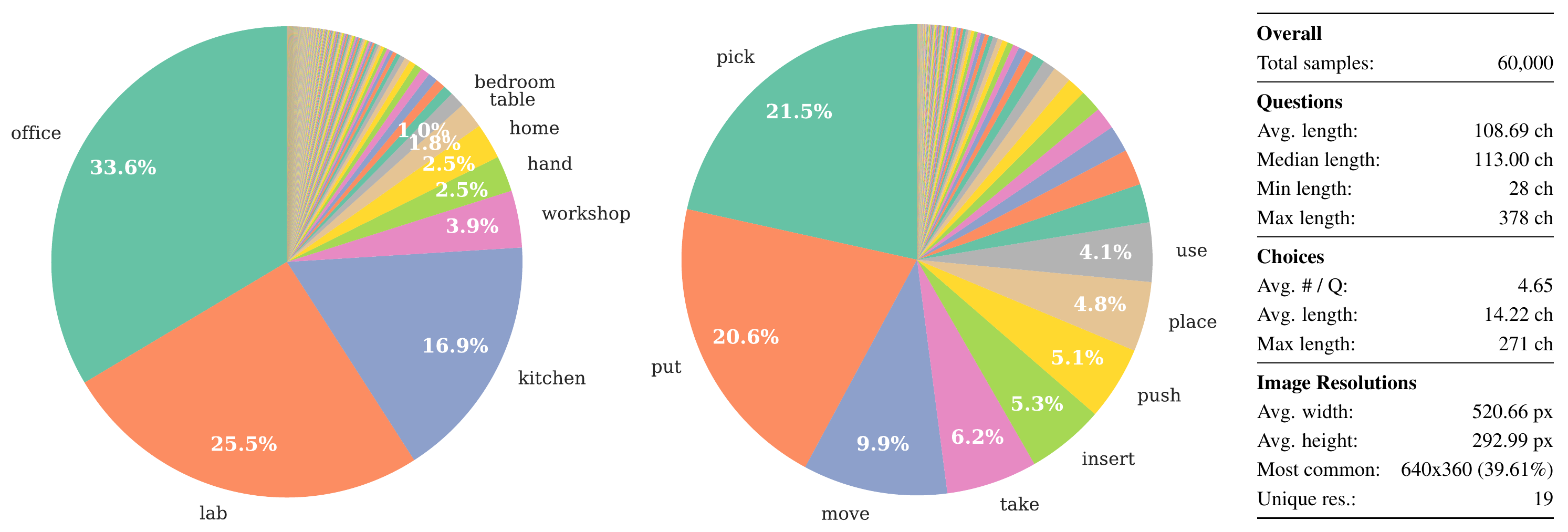}
    \caption{\textbf{Distribution and key statistics of \dsname{} dataset.} (Left) \dsname covers diverse scenes with the most frequent scenes in office (33.6\%), lab (25.3\%), and kitchen (16.9\%). (Middle) \dsname covers tasks including common manipulation actions include pick (21.5\%), put (20.6\%), and move (9.9\%). (Right) The table summarizes key dataset statistics including question characteristics, answer choices, and image resolutions.}
    \label{fig:ds:stats}
\end{figure}

\begin{wraptable}{r}{0.5\columnwidth}
  \centering
  \scriptsize
  \vspace{-2em}
  \caption{Trajectories and sensing modalities across datasets with a total of 176k trajectories. \textbf{\# Traj}: number of trajectories;
\textbf{Prop}: joint-state proprioception;
\textbf{Dpth}: depth images;
\textbf{GripAp}: gripper-aperture signal;
\textbf{\# VQA}: number of questions.
\cmark\ denotes modality is available, \xmark\ denotes  absent. }
  \label{tab:rlds_modalities}
  \resizebox{0.5\columnwidth}{!}{
\begin{tabular}{lrrrrrr}
\toprule
Dataset & \# Traj & Prop & Dpth & GripAp & \# VQA \\
\midrule
DROID~\cite{khazatsky2024droid}                         & 92k   & \cmark & \cmark & \cmark & 299k  \\
Fractal~\cite{brohan2023rt1roboticstransformerrealworld} & 73k   & \cmark & \xmark & \cmark & 267k  \\
Kuka MM~\cite{lee2019icra}                             & 3k    & \cmark & \cmark & \cmark & 25k \\
Autolab~\cite{BerkeleyUR5Website}                  & 896   & \cmark & \cmark & \cmark & 22k   \\
Sirius~\cite{liu2022robot}                             & 600   & \cmark & \xmark & \cmark & 21k \\
MVP~\cite{Radosavovic2022}                             & 480   & \cmark & \xmark & \cmark & 8k \\
VINN~\cite{pari2021surprising}                         & 435   &  \xmark  & \xmark & \xmark & 34   \\
Fanuc~\cite{fanuc_manipulation2023}                    & 415   & \cmark & \xmark & \cmark & 11k \\
TableTop~\cite{zhou2023learning}    & 110   & \cmark & \xmark & \cmark & 5k \\
VIOLA~\cite{zhu2022viola}                              & 135   & \cmark & \xmark & \cmark & 8k \\
BUDS~\cite{zhu2022bottom}                              & 50    & \cmark & \xmark & \cmark & 6k \\
ROT~\cite{haldar2023watch}                             & 14    & \cmark & \xmark & \cmark & 245   \\
\bottomrule
\end{tabular}
}
\vspace{-2em}
\end{wraptable}

\paragraph{Open X-Embodiment and its datasets} Open X-Embodiment~\cite{open_x_embodiment_rt_x_2023} is major collaborative research initiative that aggregates robotic demonstration data collected from 22 different robot embodiments across 35 research labs worldwide, encompassing over 1 million trajectories covering more than 500 skills. 
Applying domain reweighting ~\cite{xie2023doremi}, we select a subset focusing on manipulation with real robot embodiments. 
In total, we use 13 datasets~\cite{khazatsky2024droid, brohan2023rt2visionlanguageactionmodelstransfer, rosete2022tacorl, mees23hulc2, lee2019icra, BerkeleyUR5Website, liu2022robot, Radosavovic2022, fanuc_manipulation2023, zhou2023modularity, zhou2023learning, zhu2022viola, zhu2022bottom, pari2021surprising, haldar2023watch} with a total of 176,139 trajectories. 
While most modalities are included in Open X-Embodiments release, we manually include modalities introduced by the original paper. For example, DROID dataset~\cite{khazatsky2024droid} includes camera calibration information and stereo depth. The detailed modality inclusion can be found in Table. \ref{tab:rlds_modalities}.

\paragraph{\benchname for Open X-Embodiment}
We use {\benchname} to process each robot trajectory from the Open X-Embodiment dataset by selecting and interpreting the scenes.
The entire process takes 2935.7 GPU hours on Nvidia A100 GPUs.  For each selected keyframe, {\benchname} instantiates questions from embodied question templates resulting in the generation of a pool of over 3 million VQA items.


\paragraph{\dsname{} Curation}  
Inspired by data optimization paradigms such as domain reweighting in natural language processing~\cite{xie2023doremi} and robot policy learning~\cite{pmlr-v270-hejna25a}, our curation process aims to balance the distribution of questions across diverse scene and task types. It selects a representative and high-quality subset of questions that effectively balances diversity across scenes, tasks, skills, and reasoning types, while ensuring clarity and unambiguous ground truth. In total, \dsname contains 684,710 questions, spanning 463 distinct real-world scenes, 3,396 unique robotic manipulation tasks, and 149 different manipulation skills.

\begin{table*} 
    \centering
    \caption{Performance Comparison of Multimodal Foundation Models on OpenX-VQA Benchmark Categories (\%). Upper part: zero-shot. Lower part: with CoT prompting.   }
    \captionsetup{font=footnotesize} 
    \footnotesize 
    \setlength{\tabcolsep}{2.5pt} 

    \sisetup{
        table-format=2.2, 
        table-space-text-post={\%}, 
        output-decimal-marker={.}, 
        table-parse-only, 
        input-symbols = {-- xx} 
    }
\resizebox{\columnwidth}{!}{
    \begin{tabular}{@{}l
                    S[table-format=2.2, table-parse-only] 
                    S[table-format=2.2, table-parse-only] 
                    S[table-format=2.2, table-parse-only] 
                    S[table-format=2.2, table-parse-only] 
                    S[table-format=2.2, table-parse-only] 
                    S[table-format=2.2, table-parse-only] 
                    S[table-format=2.2, table-parse-only] 
                    S[table-format=2.2, table-parse-only] 
                    S[table-format=2.2, table-parse-only] 
                    S[table-format=2.2, table-parse-only] 
                    S[table-format=2.2, table-parse-only] 
                    S[table-format=2.2, table-parse-only] 
                    @{}} 

\toprule
\multicolumn{2}{c}{} 
& \multicolumn{5}{c}{\cellcolor{spatclr}\textbf{Spatial Reasoning}}
& \multicolumn{3}{c}{\cellcolor{goalclr}\textbf{Goal Reasoning}}
& \multicolumn{3}{c}{\cellcolor{intclr}\textbf{Interaction Reasoning}}\\
\cmidrule(lr){3-7}\cmidrule(lr){8-10}\cmidrule(l){11-13}
        \textbf{Model} & {\textbf{Overall}} & {\textbf{RS}} & {\textbf{OS}} & {\textbf{SR}} & {\textbf{SU}} & {\textbf{MV}} & {\textbf{TS-G}} & {\textbf{TS-S}} & {\textbf{TS-GL}} & {\textbf{AU}} & {\textbf{IP}} & {\textbf{TU}} \\
         & {(\%)} & {(\%)} & {(\%)} & {(\%)} & {(\%)} & {(\%)} & {(\%)} & {(\%)} & {(\%)} & {(\%)} & {(\%)} & {(\%)} \\ 
        \midrule
        \multicolumn{13}{@{}l}{\textit{Zero-Shot}} \\
         LLaVA 1.5-7B            & 21.58 & 35.32 & 23.87 & 16.08 & 17.78 & 17.50 & 31.82 & 23.79 & 19.03 & 20.30 & 21.74 & 22.37 \\
        LLaVA 1.6 Mistral-7B    & 24.09 & 30.31 & 35.13 & 19.42 & 20.24 & \textbf{19.29} & 34.20 & 30.77 & \textbf{19.52} & 18.67 & 20.70 & 22.83 \\
        LLaVA 1.6-34B           & 24.94 & 26.66 & 29.75 & 21.47 & 23.18 & 17.86 & 29.19 & 29.40 & 17.90 & 19.49 & 36.98 & 30.59 \\
        Llama 3.2-90B  & 28.60 & 31.94 & 55.87 & 18.51 & 26.61 & 16.43 & 28.23 & 35.27 &  8.06 & 18.13 & 51.56 & 49.77 \\

        Qwen 2.5 VL-7B    & 30.63 & 41.68 & 55.63 & 21.55 & 24.38 & 17.32 & 33.01 & 42.57 &  7.82 & 25.71 & 46.61 & 39.73 \\
        Qwen 2.5 VL-32B   & 37.68 & \textbf{49.39} & 71.37 & 21.85 & \textbf{28.53} & 17.50 & \textbf{34.21} & \textbf{55.08} & 12.90 & 30.45 & 63.80 & 49.32 \\
        Qwen 2.5 VL-72B   & \textbf{37.76} & 38.84 & \textbf{85.00} & \textbf{22.31} & 28.23 & 15.71 & 28.47 & 51.89 & 10.08 & \textbf{33.96} & \textbf{71.09} & \textbf{54.79} \\
        
         \midrule
         \multicolumn{13}{@{}l}{\textit{CoT Reasoning}} \\
          LLaVA 1.5-7B            & 21.61 & 28.28 & 21.00 & 17.37 & 20.90 & 18.93 & 25.36 & 24.19 & \textbf{21.53} & 21.24 & 20.31 & 20.09 \\
        LLaVA 1.6 Mistral-7B    & 24.05 & 27.60 & 38.87 & 17.15 & 20.18 & \textbf{22.32} & 25.84 & 28.03 & 18.47 & 18.40 & 30.60 & 29.68 \\
        LLaVA 1.6-34B           & 23.49 & 20.43 & 31.00 & 21.24 & 22.88 & 20.36 & 18.18 & 26.14 & 16.77 & 21.79 & 35.16 & 26.94 \\
        Llama 3.2-90B  & 30.45 & 32.34 & 79.87 & 13.35 & 26.37 & 18.57 & 29.90 & 29.14 & 14.27 & 19.76 & 59.24 & 44.75\\
        Qwen 2.5 VL-7B    & 34.82 & 38.02 & 90.00 & \textbf{21.78} & 23.30 & 16.79 & \textbf{36.84} & 46.48 & 18.39 & 28.15 & 42.71 & 36.99 \\
        Qwen 2.5 VL-32B & \textbf{41.30} & \textbf{48.85} & 90.50 & 18.82 & 29.19 & 19.82 & 35.17 & \textbf{60.43} & 18.71 & 32.21 & 71.35 & \textbf{49.32} \\
        Qwen 2.5 VL-72B   & 39.52 & 44.79 & \textbf{92.37} & 18.36 & \textbf{29.73} & 13.39 & 29.19 & 55.28 & 13.15 & \textbf{36.13} & \textbf{74.09} & 46.12 \\

        \bottomrule
    \end{tabular}
    }
    \captionsetup{font=scriptsize, justification=justified} 
    \caption*{
        \textit{Category Abbreviations:} 
        \textbf{Spatial Reasoning:} RS: Robot State (gripper/arm position estimation), 
        OS: Object State (object reachability/manipulability), 
        SR: Spatial Relationship (relative positioning between robot and objects), 
        SU: Scene Understanding (spatial layout comprehension), 
        MV: Multiple View (cross-view correspondence).
        \textbf{Goal-Conditioned Reasoning:} 
        TS-G: Task State-grasp (grasp stability assessment), 
        TS-S: Task State-success (task completion status), 
        TS-GL: Task State-goal (goal configuration understanding), 
        \textbf{Interaction Reasoning:} 
        AU: Action Understanding (robot's current action phase), 
        IP: Interaction Phase (prediction of next robot action), 
        TU: Trajectory Understanding (overall task interpretation).
    }
    \label{tab:updated-model-performance-neurips} 
\end{table*}

\section{Experiment}
In this section, we sample 60k VQA from \dsname with a 50k training set (\dsname-Train) and a 10k testing set (\dsname-Test). We mainly study two research questions: (1) How does \dsname-Train dataset improve the spatial and interaction reasoning capabilities of VLMs? and (2) How effectively does \dsname-Test evaluate VLMs in these reasoning tasks? 

\textbf{Evaluation Setup} We benchmark state-of-the-art open-source models in different configurations, including LLaVA, Llama 3.2 Vision, and Qwen2-VL/Qwen2.5-VL. 
Each model is evaluated under both zero-shot and Chain-of-Thought (CoT) prompting settings. For CoT, we follow the prompting strategy from \cite{geminiroboticsteam2025geminiroboticsbringingai} by appending the following instruction to the end of each question: \textit{“Reason step by step about the answer, and show your work, for each step. Only after that, proceed to the final answer."} We run a simultaneous Llama-3.2-3B-Instruct to extract model outputs for final letter answer. We focus fine-tuning on language layers (both attention and MLP modules) while keeping vision layers frozen. For each configuration, we use random 2000 questions from the testing set. For consistency, all models are evaluated with a temperature of 0.7, a maximum completion token length of 4096, and overall context length of 10240. All models use their vision or vision instruct version with float16 quantization.  All models are evaluated with 8 Nvidia A100 GPUs with 80GB memory. We use LoRA to fine-tune LLaVA 1.6 with rank 128 and alpha 256.


\subsection{Benchmark with \dsname}
Table \ref{tab:updated-model-performance-neurips} presents a detailed comparison of vision–language foundation models on the \dsname benchmark, evaluated under both zero-shot and Chain-of-Thought (CoT) prompting conditions. The results reveal nuanced interactions across model architecture, scale, and reasoning strategy.


\paragraph{Cross-Model Performance:}
Evaluation data on \dsname-test suggests that Qwen models has higher overall accuracy compared to other VLMs of the same configuration, which align with the observation from other VQA benchmarks such as ~\cite{lu2024mathvistaevaluatingmathematicalreasoning,hendrycks2021measuringmassivemultitasklanguage}. Qwen 2.5 VL-72B achieves the highest zero-shot accuracy at 37.76\%, while Qwen 2.5 VL-32B achieves 41.30\% overall accuracy in the CoT setting. Qwen models particularly excel in object-centric categories such as Object State, where Qwen 2.5 VL-72B reaches 85.00\% (zero-shot) and 92.37\% (CoT), and Interaction Phase (IP) (71.09\% zero-shot, 74.09\% CoT for 72B).

\paragraph{Impact of Model Scale.}
Zero-shot accuracy generally improves with model size — rising from 30.63\% (Qwen 7B) to 37.76\% (Qwen 72B). However, this trend does not hold in the CoT setting, where the 32B model surpasses the 72B model (41.30\% vs. 39.52\%). The observation aligns the official technical report of Qwen2.5\cite{qwen2.5} that the mathematical and problem-solving capabilities of Qwen2.5-VL-32B are further enhanced through reinforcement learning.
LLaMA models display a different trend — while the 11B model outperforms the 90B version in zero-shot setting, the larger model benefits more under CoT prompting, suggesting that scaling may unlock latent capabilities only when paired with explicit reasoning support.

\textbf{Effectiveness of CoT Prompting:}
CoT prompting generally enhances performance for both Qwen and LLaMA models. For example, Qwen 2.5 VL-7B improves from 30.63\% to 34.82\%, and LLaMA 3.2-90B increases from 28.60\% to 30.45\%. The most substantial gains are observed in Qwen 2.5 VL-32B, which improves from 37.68\% to 41.30\%. Results suggest that CoT benefits Task State–Success(from 55.08\% to 60.43\%), and Interaction Phase (from 63.80\% to 71.35\%). However, in the Spatial Relationship category, for example, Qwen 32B's accuracy drops from 21.85\% to 18.82\%, indicating that verbose reasoning chains may introduce noise in tasks requiring precise spatial localization. 


\subsection{Finetuning with \dsname}

\begin{figure}
    \centering
    \includegraphics[width=\linewidth]{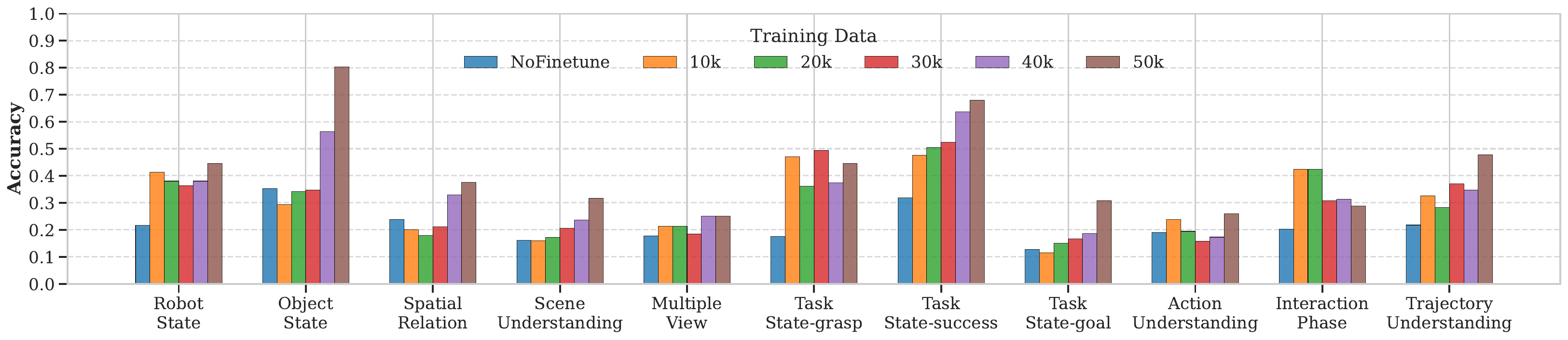}
    \caption{ \textbf{Fine-tuning LLaVA 1.6 with increasing training data of \dsname} from 10k to 50k VQA items. Accuracy improvements almost all categories compared to no fine-tuning.}
    \label{fig:finetuning}
\end{figure}
We perform model finetuning experiment using \dsname-train and evaluate on \dsname-test. We increase the training data samples from 10k to 50k in finetuning. 
As depicted in Figure~\ref{fig:finetuning}, increasing the fine-tuning data generally leads to notable performance enhancements across most VQA categories.
Significant gains are observed in `Object State' understanding, where accuracy improved from 29.34\%  to 80.24\%. ``Task State-success'' also sees a substantial rise from 47.65\% to 68.03\%. Other categories demonstrating clear positive trends with more data. However, in some categories such as Spatial Relationship and Task State–Goal, fine-tuning with limited data (e.g., 10k) underperforms the no-finetuning baseline. This may be because the model has not yet seen enough task-specific examples to begin generalizing, or because the question formats in \dsname differ from those seen during pretraining, 
requiring adaptation time. In some categories, finetuning with \dsname does not improve the performance due to the reasoning capability limitation of the base model. This is also reflected in the fact that LLaVA shows performance degradation in CoT prompting in Table \ref{tab:updated-model-performance-neurips}. The ``interaction phase'' question requires the model to predict the next frame, demanding complex reasoning and making it a particularly challenging problem. This suggests that for complex tasks, the base model language performance is important for further improvement with \dsname.

\subsection{Comparison with Human Performance}

\begin{wrapfigure}{r}{0.4\columnwidth}
    \centering
    \includegraphics[width=\linewidth]{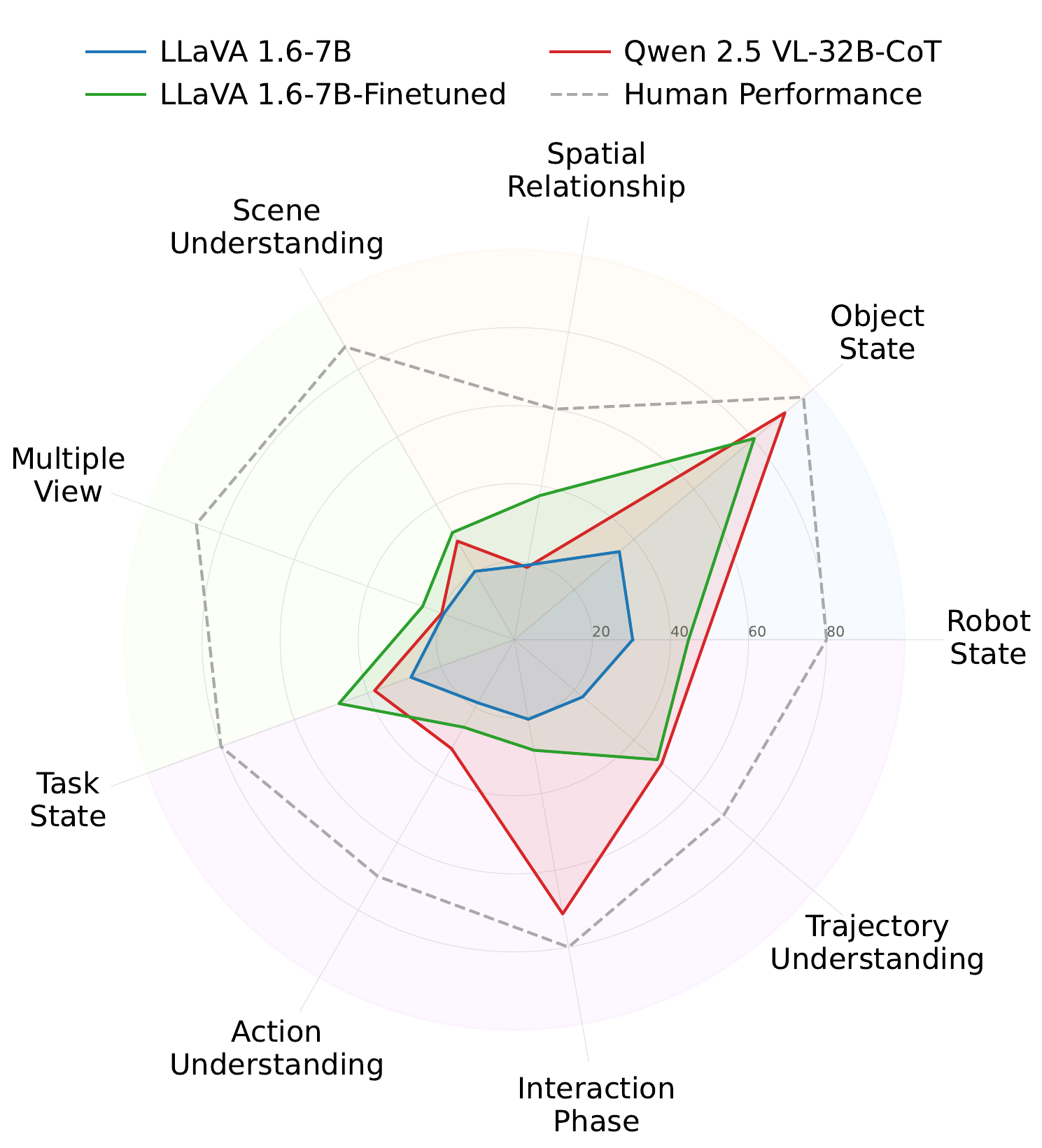}
    \caption{Comparison of human performance to different multimodal foundation models.}
    \label{fig:radar}
\end{wrapfigure}
We conducted a human evaluation covering all 11 categories defined in Table \ref{tab:updated-model-performance-neurips}. For each category, a human evaluator was asked to randomly answer questions 
from \dsname-test. We use the average success rate as a reference for comparison with three models—LLaVA 1.6-7B, LLaVA 1.6-7B-Finetuned, and Qwen 2.5 VL-32B—CoT on the same set of categories as shown in Figure~\ref{fig:radar}.
Qwen 2.5 VL-32B—CoT achieves near human accuracy, with 90.5\% in Object State compared to 96.7\% for humans, and 71.35\% in Interaction Phase versus the human score of 80.0\%. In more complex spatial reasoning tasks such as Spatial Relationship, where human achieves 60.0\% accuracy, the best model (LLaVa 1.6-7B, finetuned) reaches only 19.42\%. 
This may suggest that even if observing from multiple views, a monocular image may lack the full depth information needed to accurately determine the spatial relationship.
Furthermore, finetuning enhances model performance. LLaVA 1.6-7B finetuned on the Robo2VLM-1 training dataset shows consistent improvements across multiple categories, particularly in Task State, Object State, and Trajectory Understanding, compared to its non-finetuned LLaVA 1.6-7B. These findings demonstrate the potential \dsname in studying and narrowing the gap between model and human performance in spatial and task reasoning.

\section{Conclusion and Discussion}

In this paper, we introduce \benchname, a framework that generates VQA grounded in robot sensory modalities. We apply \benchname to 176k real robot trajectories from Open X-Embodiment, and curate \dsname, a comprehensive dataset of 684,710 questions covering 463 distinct scenes, 3,396 robotic manipulation tasks, and 149 manipulation skills.
Evaluation of state-of-the-art open-source VLMs suggests that some VLMs, such as Qwen2.5 VL 32B with CoT prompting, can achieve near human performance in questions related to object reachability and interaction understanding, while there is a significant gap to human in reasoning fine-grained spatial relationship and interactions. Evaluation also suggests that finetuning  \dsname dataset improves in spatial and interaction reasoning. Future work will focus on generalizing \benchname to a wider range of robot embodiments and generating an even more diverse dataset. We also plan to explore the deployment of models trained on \dsname to real-world robotic tasks.

\textbf{Limitation} We acknowledge that \benchname is a data generation framework that relies on the quality of input tele-operated trajectories. If the original trajectory is wrongly calibrated, it compromises the quality of generated VQA data. Or if the original trajectory misses embodiment sensory modalities, such as NYU VINN~\cite{pari2021surprising} (0.2\% of the 176k trajectories), it limits the amount of questions that \benchname can generate. 

\section*{Acknowledgement}
This research was performed at the AUTOLAB at UC Berkeley in affiliation with the Berkeley AI Research (BAIR) Lab. This work is supported in part by donations from Google.


\bibliographystyle{unsrt}
\bibliography{syxie}

\begin{thebibliography}{10}

\bibitem{radford2021learning}
Alec Radford, Jong~Wook Kim, Chris Hallacy, Aditya Ramesh, Gabriel Goh, Sandhini Agarwal, Girish Sastry, Amanda Askell, Pamela Mishkin, Jack Clark, Gretchen Krueger, and Ilya Sutskever.
\newblock Learning transferable visual models from natural language supervision.
\newblock In Marina Meila and Tong Zhang, editors, {\em Proceedings of the 38th International Conference on Machine Learning}, volume 139 of {\em Proceedings of Machine Learning Research}, pages 8748--8763. PMLR, 18--24 Jul 2021.

\bibitem{qwen2.5}
Qwen Team.
\newblock Qwen2.5: A party of foundation models, September 2024.

\bibitem{touvron2023Llama2}
Hugo Touvron, Louis Martin, Kevin Stone, Peter Albert, Amjad Almahairi, Yasmine Babaei, Nikolay Bashlykov, Soumya Batra, Prajjwal Bhargava, Shruti Bhosale, et~al.
\newblock Llama 2: Open foundation and fine-tuned chat models.
\newblock {\em arXiv preprint arXiv:2307.09288}, 2023.

\bibitem{liu2023visual}
Haotian Liu, Chunyuan Li, Qingyang Wu, and Yong~Jae Lee.
\newblock Visual instruction tuning.
\newblock {\em Advances in neural information processing systems}, 36:34892--34916, 2023.

\bibitem{anthropic2024claude35}
Anthropic.
\newblock {Claude 3.5 Sonnet}.
\newblock \url{https://www.anthropic.com/news/claude-3-5-sonnet}, June 2024.

\bibitem{openai2024gpt4o}
OpenAI.
\newblock {GPT-4o System Card}.
\newblock \url{https://openai.com/index/gpt-4o-system-card/}, August 2024.

\bibitem{google2025gemini25}
Koray Kavukcuoglu.
\newblock {Gemini 2.5: Our most intelligent AI model}.
\newblock \url{https://blog.google/technology/google-deepmind/gemini-model-thinking-updates-march-2025/}, March 2025.

\bibitem{Chen_2024_CVPR}
Boyuan Chen, Zhuo Xu, Sean Kirmani, Brain Ichter, Dorsa Sadigh, Leonidas Guibas, and Fei Xia.
\newblock Spatialvlm: Endowing vision-language models with spatial reasoning capabilities.
\newblock In {\em Proceedings of the IEEE/CVF Conference on Computer Vision and Pattern Recognition (CVPR)}, pages 14455--14465, June 2024.

\bibitem{karamcheti2024prismatic}
Siddharth Karamcheti, Suraj Nair, Ashwin Balakrishna, Percy Liang, Thomas Kollar, and Dorsa Sadigh.
\newblock Prismatic vlms: Investigating the design space of visually-conditioned language models, 2024.

\bibitem{kim24openvla}
{Moo Jin} Kim, Karl Pertsch, Siddharth Karamcheti, Ted Xiao, Ashwin Balakrishna, Suraj Nair, Rafael Rafailov, Ethan Foster, Grace Lam, Pannag Sanketi, Quan Vuong, Thomas Kollar, Benjamin Burchfiel, Russ Tedrake, Dorsa Sadigh, Sergey Levine, Percy Liang, and Chelsea Finn.
\newblock Openvla: An open-source vision-language-action model.
\newblock {\em arXiv preprint arXiv:2406.09246}, 2024.

\bibitem{geminiroboticsteam2025geminiroboticsbringingai}
Gemini~Robotics Team, Saminda Abeyruwan, et~al.
\newblock Gemini robotics: Bringing ai into the physical world, 2025.

\bibitem{shi2025hirobotopenendedinstruction}
Lucy~Xiaoyang Shi, Brian Ichter, Michael Equi, Liyiming Ke, Karl Pertsch, Quan Vuong, James Tanner, Anna Walling, Haohuan Wang, Niccolo Fusai, Adrian Li-Bell, Danny Driess, Lachy Groom, Sergey Levine, and Chelsea Finn.
\newblock Hi robot: Open-ended instruction following with hierarchical vision-language-action models, 2025.

\bibitem{islam2024eqamx}
Md Mofijul Islam, Alexi Gladstone, Riashat Islam, and Tariq Iqbal.
\newblock {EQA-MX}: Embodied question answering using multimodal expression.
\newblock In {\em Proc.\ International Conference on Learning Representations (ICLR)}, 2024.

\bibitem{yang2025embodiedbench}
Rui Yang, Hanyang Chen, Junyu Zhang, Mark Zhao, Cheng Qian, Kangrui Wang, Qineng Wang, Teja Venkat Koripella, Marziyeh Movahedi, Manling Li, Heng Ji, Huan Zhang, and Tong Zhang.
\newblock {EMBODIEDBENCH}: Comprehensive benchmarking multi-modal large language models for vision-driven embodied agents.
\newblock {\em arXiv preprint arXiv:2502.09560}, 2025.

\bibitem{li2024eai}
Manling Li, Shiyu Zhao, Qineng Wang, Kangrui Wang, Yu Zhou, Sanjana Srivastava, Cem Gokmen, Tony Lee, Li Erran Li, Ruohan Zhang, Weiyu Liu, Percy Liang, Li Fei-Fei, Jiayuan Mao, and Jiajun Wu.
\newblock Embodied agent interface: Benchmarking {LLMs} for embodied decision making.
\newblock In {\em NeurIPS 2024 Track on Datasets and Benchmarks}, 2024.

\bibitem{shridhar2020alfred}
Mohit Shridhar, Jesse Thomason, Daniel Gordon, Yonatan Bisk, Winson Han, Roozbeh Mottaghi, Luke Zettlemoyer, and Dieter Fox.
\newblock {ALFRED}: A benchmark for interpreting grounded instructions for household robots.
\newblock In {\em Proceedings of the IEEE/CVF Conference on Computer Vision and Pattern Recognition (CVPR)}, 2020.

\bibitem{szot2021habitat}
Andrew Szot, Edward Coumans, Alex Collett, and et al.
\newblock Habitat 2.0: Training home assistants to rearrange their habitat.
\newblock In {\em Proceedings of the International Conference on Neural Information Processing Systems (NeurIPS)}, 2021.

\bibitem{kolve2017ai2thor}
Eric Kolve, Roozbeh Mottaghi, Daniel Gordon, and et al.
\newblock {AI2‑THOR}: An interactive 3d environment for visual {AI}.
\newblock {\em arXiv preprint arXiv:1712.05474}, 2017.

\bibitem{sermanet2023robovqa}
Pierre Sermanet, Tianli Ding, Jeffrey Zhao, Fei Xia, Debidatta Dwibedi, Keerthana Gopalakrishnan, Christine Chan, Gabriel Dulac-Arnold, Sharath Maddineni, Nikhil J. Joshi, Pete Florence, Wei Han, Robert Baruch, Yao Lu, Suvir Mirchandani, Peng Xu, Pannag Sanketi, Karol Hausman, Izhak Shafran, Brian Ichter, and Yuan Cao.
\newblock Robovqa: Multimodal long-horizon reasoning for robotics.
\newblock {\em arXiv preprint arXiv:2311.00899}, 2023.

\bibitem{ji2025robobrain}
Yuheng Ji, Huajie Tan, Jiayu Shi, Xiaoshuai Hao, Yuan Zhang, Hengyuan Zhang, Pengwei Wang, Mengdi Zhao, Yao Mu, Pengju An, et~al.
\newblock Robobrain: A unified brain model for robotic manipulation from abstract to concrete.
\newblock {\em CVPR}, 2025.

\bibitem{levine2016end}
Sergey Levine, Chelsea Finn, Trevor Darrell, and Pieter Abbeel.
\newblock End-to-end training of deep visuomotor policies.
\newblock {\em Journal of Machine Learning Research}, 17(39):1--40, 2016.

\bibitem{octo_2023}
{Octo Model Team}, Dibya Ghosh, Homer Walke, Karl Pertsch, Kevin Black, Oier Mees, Sudeep Dasari, Joey Hejna, Charles Xu, Jianlan Luo, Tobias Kreiman, {You Liang} Tan, Dorsa Sadigh, Chelsea Finn, and Sergey Levine.
\newblock Octo: An open-source generalist robot policy.
\newblock \url{https://octo-models.github.io}, 2023.

\bibitem{chi2023diffusionpolicy}
Cheng Chi, Siyuan Feng, Yilun Du, Zhenjia Xu, Eric Cousineau, Benjamin Burchfiel, and Shuran Song.
\newblock Diffusion policy: Visuomotor policy learning via action diffusion.
\newblock In {\em Proceedings of Robotics: Science and Systems (RSS)}, 23.

\bibitem{open_x_embodiment_rt_x_2023}
Open X-Embodiment Collaboration, Abby O'Neill, Abdul Rehman, Abhinav Gupta, Abhiram Maddukuri, Abhishek Gupta, Abhishek Padalkar, et~al.
\newblock Open {X-E}mbodiment: Robotic learning datasets and {RT-X} models.
\newblock \url{https://arxiv.org/abs/2310.08864}, 2023.

\bibitem{xie2023doremi}
Sang~Michael Xie, Hieu Pham, Xuanyi Dong, Nan Du, Hanxiao Liu, Yifeng Lu, Percy~S Liang, Quoc~V Le, Tengyu Ma, and Adams~Wei Yu.
\newblock Doremi: Optimizing data mixtures speeds up language model pretraining.
\newblock {\em Advances in Neural Information Processing Systems}, 36:69798--69818, 2023.

\bibitem{pmlr-v270-hejna25a}
Joey Hejna, Chethan~Anand Bhateja, Yichen Jiang, Karl Pertsch, and Dorsa Sadigh.
\newblock Remix: Optimizing data mixtures for large scale imitation learning.
\newblock In Pulkit Agrawal, Oliver Kroemer, and Wolfram Burgard, editors, {\em Proceedings of The 8th Conference on Robot Learning}, volume 270 of {\em Proceedings of Machine Learning Research}, pages 145--164. PMLR, 06--09 Nov 2025.

\bibitem{droid}
Abhishek Sharma, Vishal Sundaresan, Yizhou Zhu, Parth Shah, Kuan Liu, Michael Laskin, Jonathan Tompson, Ayzaan Wahid, Yevgen Chebotar, and Karol Hausman.
\newblock Droid: A large-scale in-the-wild robot manipulation dataset.
\newblock {\em arXiv preprint arXiv:2310.01894}, 2023.

\bibitem{brohan2023rt1roboticstransformerrealworld}
Anthony Brohan et~al.
\newblock Rt-1: Robotics transformer for real-world control at scale.
\newblock 2023.

\bibitem{brohan2023rt2visionlanguageactionmodelstransfer}
Anthony Brohan et~al.
\newblock Rt-2: Vision-language-action models transfer web knowledge to robotic control, 2023.

\bibitem{black2024pi0}
Kevin Black, Noah Brown, Danny Driess, Adnan Esmail, Michael Equi, Chelsea Finn, Niccolo Fusai, Lachy Groom, Karol Hausman, Brian Ichter, Szymon Jakubczak, Tim Jones, Liyiming Ke, Sergey Levine, Adrian Li-Bell, Mohith Mothukuri, Suraj Nair, Karl Pertsch, Lucy~Xiaoyang Shi, James Tanner, Quan Vuong, Anna Walling, Haohuan Wang, and Ury Zhilinsky.
\newblock $\pi_0$: A vision-language-action flow model for general robot control.
\newblock \url{https://physicalintelligence.company/blog/pi0}, 2024.

\bibitem{khazatsky2024droid}
Alexander Khazatsky, Karl Pertsch, et~al.
\newblock Droid: A large-scale in-the-wild robot manipulation dataset.
\newblock 2024.

\bibitem{rosete2022tacorl}
Erick Rosete-Beas, Oier Mees, Gabriel Kalweit, Joschka Boedecker, and Wolfram Burgard.
\newblock Latent plans for task agnostic offline reinforcement learning.
\newblock 2022.

\bibitem{mees23hulc2}
Oier Mees, Jessica Borja-Diaz, and Wolfram Burgard.
\newblock Grounding language with visual affordances over unstructured data.
\newblock In {\em Proceedings of the IEEE International Conference on Robotics and Automation (ICRA)}, London, UK, 2023.

\bibitem{lee2019icra}
Michelle~A Lee, Yuke Zhu, Krishnan Srinivasan, Parth Shah, Silvio Savarese, Li~Fei-Fei, Animesh Garg, and Jeannette Bohg.
\newblock Making sense of vision and touch: Self-supervised learning of multimodal representations for contact-rich tasks.
\newblock In {\em 2019 IEEE International Conference on Robotics and Automation (ICRA)}, 2019.

\bibitem{BerkeleyUR5Website}
Lawrence~Yunliang Chen, Simeon Adebola, and Ken Goldberg.
\newblock Berkeley {UR5} demonstration dataset.
\newblock https://sites.google.com/view/berkeley-ur5/home.

\bibitem{liu2022robot}
Huihan Liu, Soroush Nasiriany, Lance Zhang, Zhiyao Bao, and Yuke Zhu.
\newblock Robot learning on the job: Human-in-the-loop autonomy and learning during deployment.
\newblock In {\em Robotics: Science and Systems (RSS)}, 2023.

\bibitem{Radosavovic2022}
Ilija Radosavovic, Tete Xiao, Stephen James, Pieter Abbeel, Jitendra Malik, and Trevor Darrell.
\newblock Real-world robot learning with masked visual pre-training.
\newblock In {\em CoRL}, 2022.

\bibitem{pari2021surprising}
Jyothish Pari, Nur~Muhammad Shafiullah, Sridhar~Pandian Arunachalam, and Lerrel Pinto.
\newblock The surprising effectiveness of representation learning for visual imitation, 2021.

\bibitem{fanuc_manipulation2023}
Xinghao Zhu, Ran Tian, Chenfeng Xu, Mingyu Ding, Wei Zhan, and Masayoshi Tomizuka.
\newblock Fanuc manipulation: A dataset for learning-based manipulation with fanuc mate 200id robot.
\newblock 2023.

\bibitem{zhou2023modularity}
Yifan Zhou, Shubham Sonawani, Mariano Phielipp, Simon Stepputtis, and Heni Amor.
\newblock Modularity through attention: Efficient training and transfer of language-conditioned policies for robot manipulation.
\newblock In {\em Conference on Robot Learning}, pages 1684--1695. PMLR, 2023.

\bibitem{zhou2023learning}
Yifan Zhou, Shubham Sonawani, Mariano Phielipp, Heni Ben~Amor, and Simon Stepputtis.
\newblock Learning modular language-conditioned robot policies through attention.
\newblock {\em Autonomous Robots}, pages 1--21, 2023.

\bibitem{zhu2022viola}
Yifeng Zhu, Abhishek Joshi, Peter Stone, and Yuke Zhu.
\newblock Viola: Imitation learning for vision-based manipulation with object proposal priors.
\newblock {\em 6th Annual Conference on Robot Learning (CoRL)}, 2022.

\bibitem{zhu2022bottom}
Yifeng Zhu, Peter Stone, and Yuke Zhu.
\newblock Bottom-up skill discovery from unsegmented demonstrations for long-horizon robot manipulation.
\newblock {\em IEEE Robotics and Automation Letters}, 7(2):4126--4133, 2022.

\bibitem{haldar2023watch}
Siddhant Haldar, Vaibhav Mathur, Denis Yarats, and Lerrel Pinto.
\newblock Watch and match: Supercharging imitation with regularized optimal transport.
\newblock In {\em Conference on Robot Learning}, pages 32--43. PMLR, 2023.

\bibitem{antol2015vqa}
Stanislaw Antol, Aishwarya Agrawal, Jiasen Lu, Margaret Mitchell, Dhruv Batra, C~Lawrence Zitnick, and Devi Parikh.
\newblock Vqa: Visual question answering.
\newblock In {\em Proceedings of the IEEE international conference on computer vision}, pages 2425--2433, 2015.

\bibitem{goyal2017making}
Yash Goyal, Tejas Khot, Douglas Summers-Stay, Dhruv Batra, and Devi Parikh.
\newblock Making the v in vqa matter: Elevating the role of image understanding in visual question answering.
\newblock In {\em Proceedings of the IEEE conference on computer vision and pattern recognition}, pages 6904--6913, 2017.

\bibitem{lee2022what}
Jae~Hee Lee, Matthias Kerzel, Kyra Ahrens, Cornelius Weber, and Stefan Wermter.
\newblock What is right for me is not yet right for you: A dataset for grounding relative directions via multi-task learning.
\newblock {\em arXiv preprint arXiv:2205.02671}, 2022.

\bibitem{das2018embodiedqa}
Abhishek Das, Samyak Datta, Georgia Gkioxari, Stefan Lee, Devi Parikh, and Dhruv Batra.
\newblock Embodied question answering.
\newblock In {\em Proceedings of the IEEE Conference on Computer Vision and Pattern Recognition (CVPR)}, 2018.

\bibitem{anderson2018vision}
Peter Anderson, Qi~Wu, Damien Teney, Joel Bruce, Mark Johnson, Stephen Gould, and Anton van~den Hengel.
\newblock Vision-and-language navigation: Interpreting visually-grounded navigation instructions in real environments.
\newblock In {\em Proceedings of the IEEE Conference on Computer Vision and Pattern Recognition (CVPR)}, 2018.

\bibitem{DBLP:reference/robo/ChristensenH16}
Henrik~I. Christensen and Gregory~D. Hager.
\newblock Sensing and estimation.
\newblock In Bruno Siciliano and Oussama Khatib, editors, {\em Springer Handbook of Robotics}, Springer Handbooks, pages 91--112. Springer, 2016.

\bibitem{lu2024mathvistaevaluatingmathematicalreasoning}
Pan Lu, Hritik Bansal, Tony Xia, Jiacheng Liu, Chunyuan Li, Hannaneh Hajishirzi, Hao Cheng, Kai-Wei Chang, Michel Galley, and Jianfeng Gao.
\newblock Mathvista: Evaluating mathematical reasoning of foundation models in visual contexts, 2024.

\bibitem{hendrycks2021measuringmassivemultitasklanguage}
Dan Hendrycks, Collin Burns, Steven Basart, Andy Zou, Mantas Mazeika, Dawn Song, and Jacob Steinhardt.
\newblock Measuring massive multitask language understanding, 2021.

\end{thebibliography}

\appendix

\newpage
\tableofcontents
\section{Broader Impact}
The development of \benchname and \dsname aims to accelerate progress in robotic manipulation by providing a robust framework for evaluating and improving Vision-Language Models. Positive societal impacts are significant. More capable robots, enhanced by VLMs rigorously tested on such benchmarks, can revolutionize various sectors. In manufacturing, they can lead to more efficient, flexible, and safer production lines by undertaking complex assembly or hazardous material handling. In healthcare, advanced robotic assistants could support surgeons with greater precision, provide personalized care for the elderly or individuals with disabilities, thereby improving their quality of life and independence, and assist in laboratory automation for faster medical research. For domestic tasks, robots could alleviate household burdens, freeing up human time for more creative or relational pursuits. Beyond these, such advancements can contribute to safer work environments by automating dangerous jobs in construction, mining, or disaster response, and even aid in environmental conservation efforts through automated monitoring and intervention. The increased productivity and innovation spurred by these technologies could lead to economic growth and the creation of new job categories focused on designing, maintaining, and overseeing these intelligent systems. However, it is important to consider potential negative societal impacts. As VLMs become more powerful through evaluation on such benchmarks, there's a risk of misuse if these capabilities are applied to autonomous systems without appropriate safeguards, potentially leading to unintended actions or job displacement in certain sectors. For example, if the underlying trajectory data in \benchname inadvertently contains biases (e.g., related to specific environments, objects, or human demonstrators), models trained or evaluated on \dsname might perpetuate or amplify these biases. Future work should actively consider methods to detect and mitigate such biases in the dataset and the models. Furthermore, while the goal is to advance AI for beneficial applications, any significant improvement in generative or understanding capabilities of models could, in principle, be adapted for unintended purposes. Therefore, ongoing discussion and development of ethical guidelines and safety protocols are crucial as VLM capabilities advance in robotics and other fields.

\section{Question Analysis}
The complete dataset can be found in the huggingface website, \url{https://huggingface.co/datasets/keplerccc/Robo2VLM-1}. We provide representative examples to show the diversity and quality of the dataset. Each VQA contains one/multiple images showing the robot current position and the scene, a language description question, and multiple choices as candidate answer. 

\subsection{Example Questions from Different Tasks}
\begin{figure}[H]
    \centering
    \includegraphics[width=\linewidth]{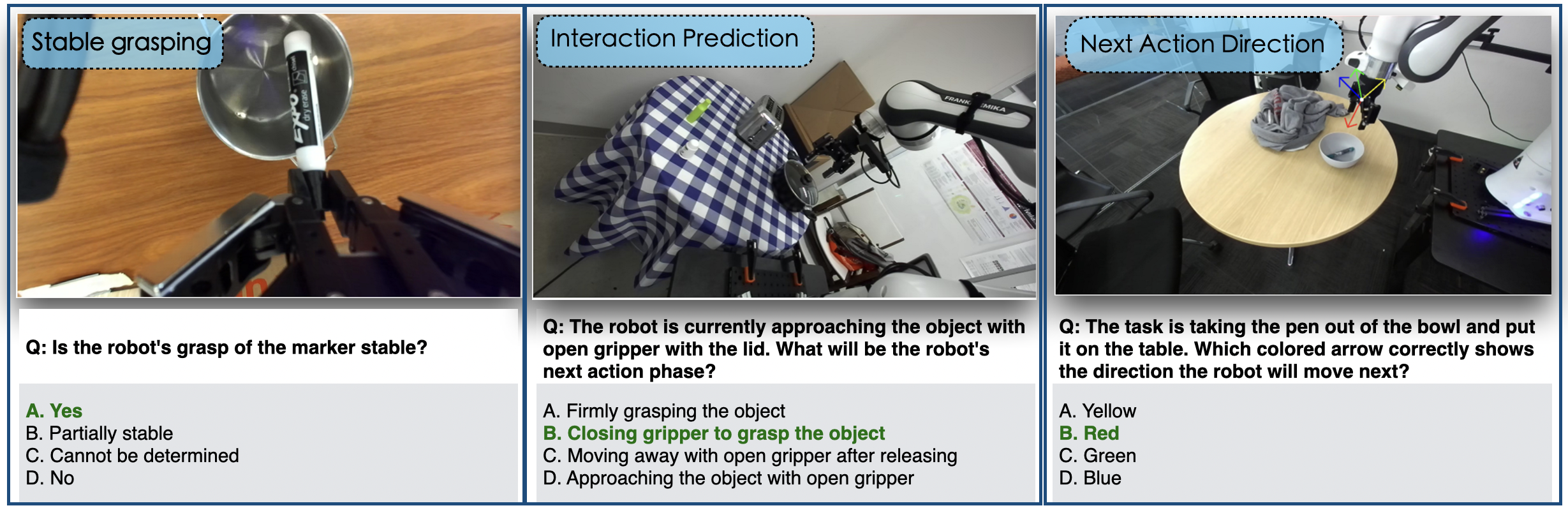}
    \caption{\textbf{Example VQAs}. Each panel illustrates a distinct category of visual question answering grounded in real robot interactions.}
    \label{fig:exp2}
\end{figure}
\begin{figure}[H]
    \centering
    \includegraphics[width=\linewidth]{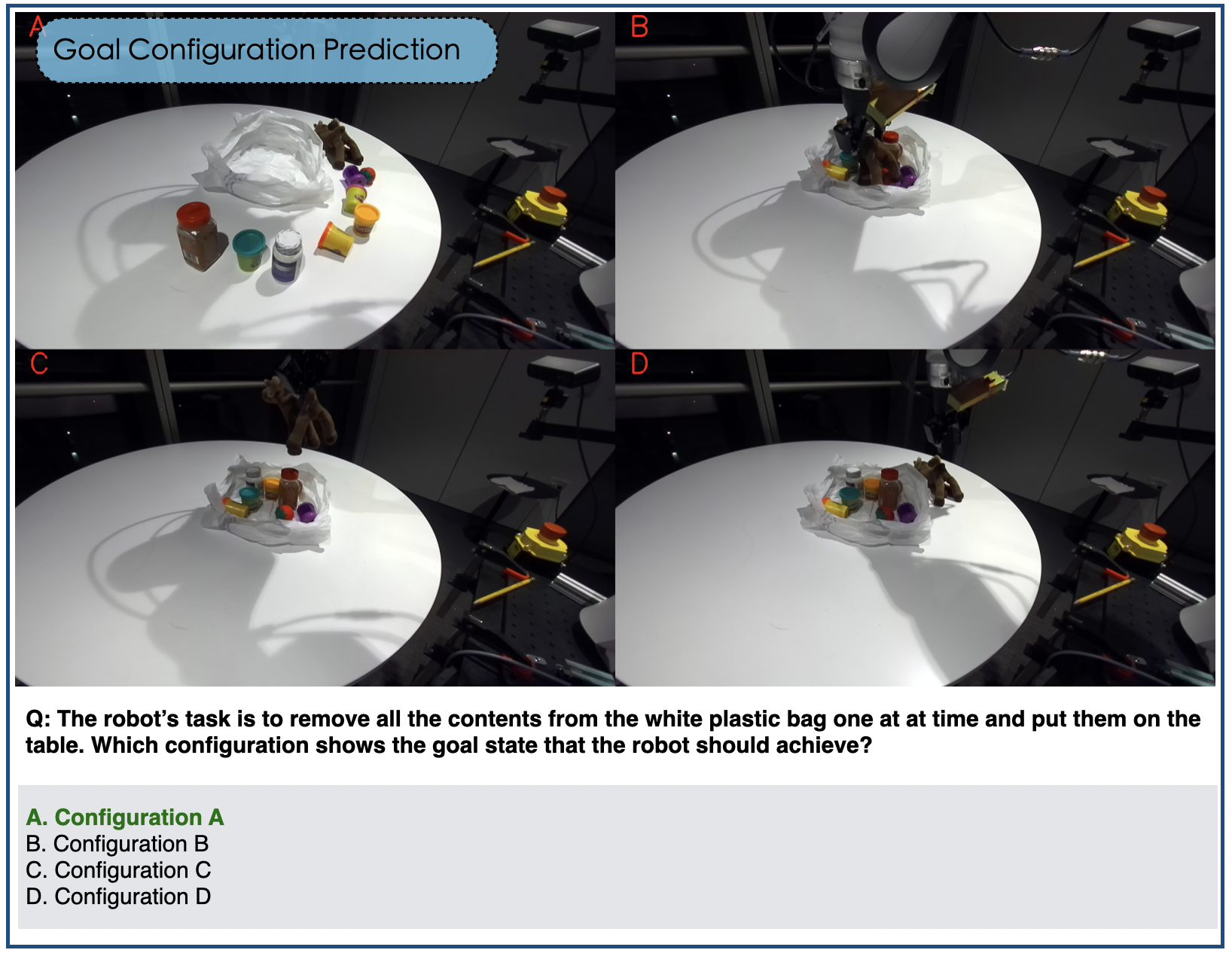}
    \includegraphics[width=\linewidth]{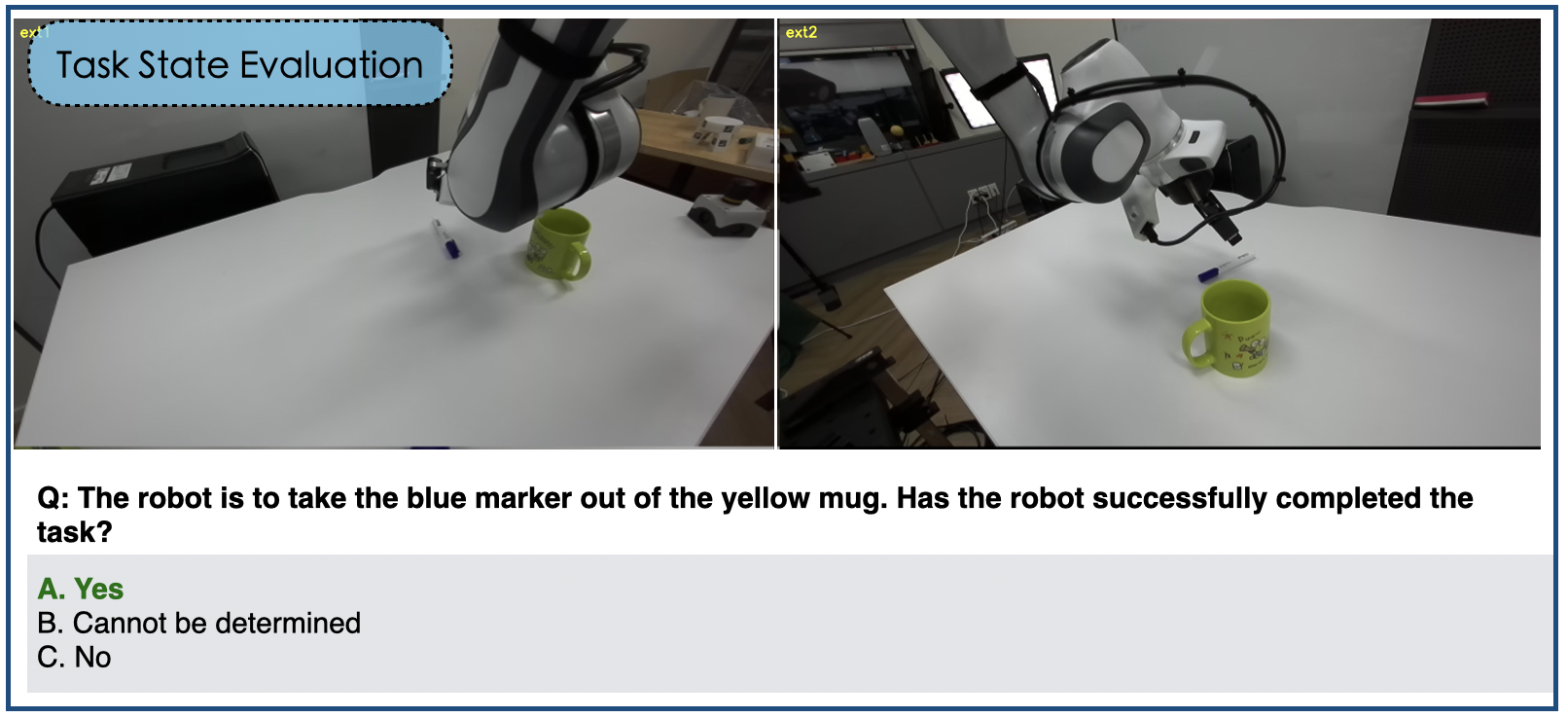}
    \caption{\textbf{Example VQAs using with mutlple images.} Each panel presents a unique type of VQA grounded in real-world robot trajectories. Goal Configuration Prediction (top) asks which scene configuration matches the task goal. Task State Evaluation (bottom) queries whether the robot has successfully completed a specified action. These examples demonstrate the need for multimodal reasoning over visual observations and task context. Correct answers are highlighted in green.}
    \label{fig:exp1}
\end{figure}

The examples in Figs.~\ref{fig:exp2},\ref{fig:exp1} highlight the diversity and complexity of visual question answering (VQA) tasks grounded in real-world robotic manipulation. Each question may be associated with multiple images, which can originate from different phases of the manipulation sequence or from distinct camera viewpoints. This design reflects the inherently temporal and multi-perspective nature of robotic tasks, requiring models to reason over a sequence of actions or fuse complementary observations. The questions span reasoning types such as goal configuration prediction, task outcome evaluation, grasp stability assessment, and interaction phase forecasting. These diverse formats challenge models to integrate spatial understanding, temporal progression, and multimodal cues, making the dataset a rigorous benchmark for evaluating the task-level reasoning capabilities of vision-language models in robotics.

\subsection{Challenging Questions}

\definecolor{questionbg}{RGB}{240,248,255}
\definecolor{answergreen}{RGB}{34,139,34}
\definecolor{reasonblue}{RGB}{65,105,225}

\newtcolorbox{questionbox}{colback=questionbg,colframe=black!50,arc=4pt,auto outer arc,left=2mm,right=2mm,top=1mm,bottom=1mm}

\newcommand{\correctanswer}[1]{\textcolor{answergreen}{\textbf{Correct Answer:} #1}}

\newcommand{\rationale}[1]{\textcolor{reasonblue}{\textbf{Expert Rationale:} #1}}

\newcommand{\vqafigure}[5]{
\begin{figure}[H]
    \centering
    \includegraphics[width=0.6\linewidth]{#1}
    \begin{questionbox}
        \textbf{Question:} #2
        \begin{itemize}[leftmargin=*]
            \item \correctanswer{#3}
            \item \rationale{#4}
        \end{itemize}
    \end{questionbox}
    \caption{#5}
    \label{fig:#1}
\end{figure}
}

\newcommand{\vqafigureduo}[5]{
\begin{figure}[H]
    \centering
    \includegraphics[width=\linewidth]{#1}
    \begin{questionbox}
        \textbf{Question:} #2
        \begin{itemize}[leftmargin=*]
            \item \correctanswer{#3}
            \item \rationale{#4}
        \end{itemize}
    \end{questionbox}
    \caption{#5}
    \label{fig:#1}
\end{figure}
}

The following figures illustrate several visual question answering (VQA) tasks conducted using robotic trajectories. Each figure presents a unique scenario where human expertise was used to validate the correctness of robotic actions or spatial understanding based on visual inspection. These are questions human experts consider challenging but answered correctly. We will introduce more details for human expert instruction and feedback in Sec.~\ref{sec:human-expert}.

\vqafigure{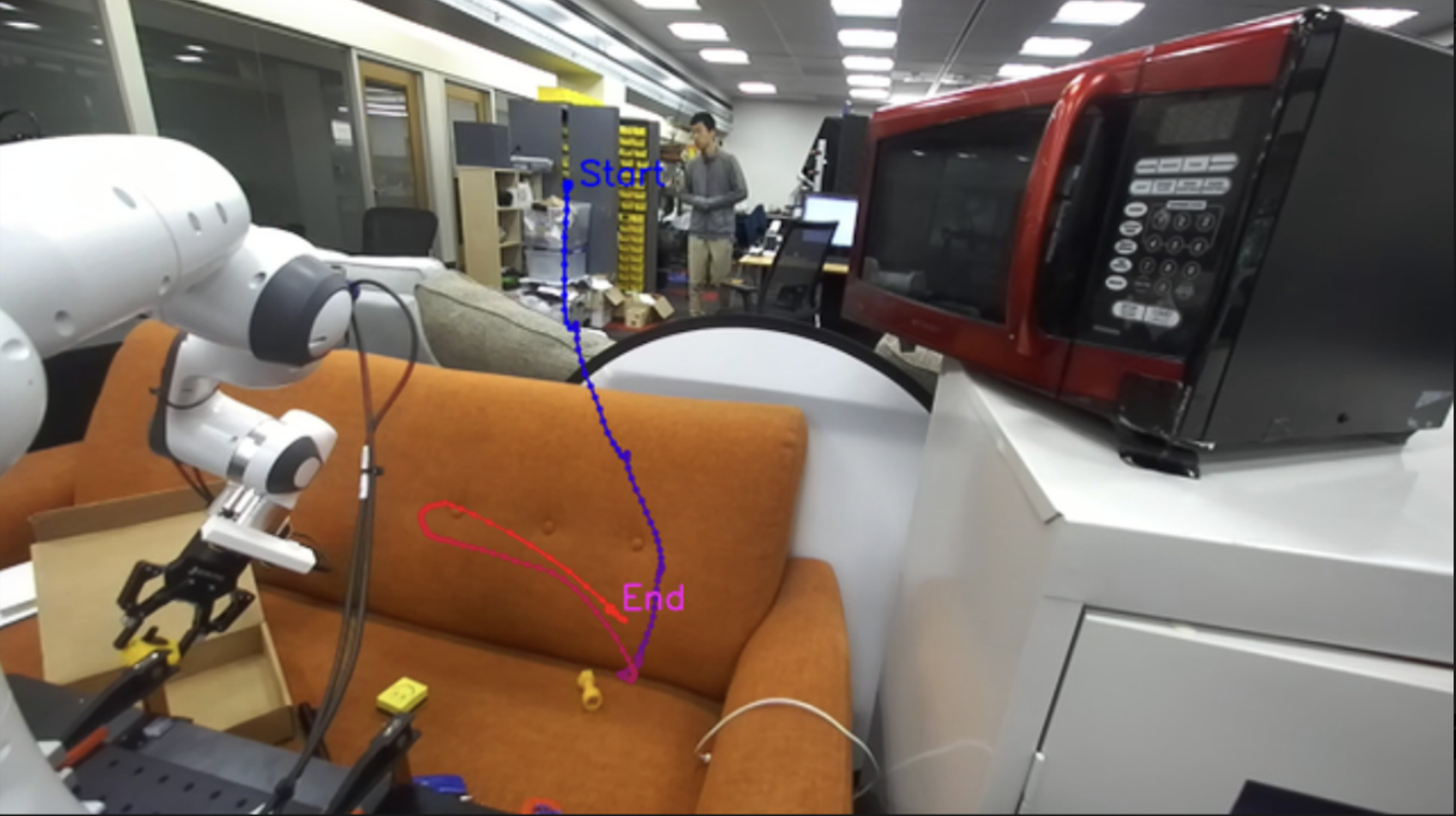}%
{Which language instruction best describes the robot's trajectory shown in the image?\\
\emph{[Pick up the black from the drawer, Drop the box into the shelf, put the yellow and black object in the box, Align the black with the table, Move the box to the floor]}}%
{Put the yellow and black object in the box}%
{The trajectory isn't directly at the objects, but the gripper position suggested interaction with the box. This reasoning led me to identify the correct choice clearly.}%
{Identifying the appropriate language instruction corresponding to a robot trajectory involving interaction with a yellow and black object.}

\vqafigure{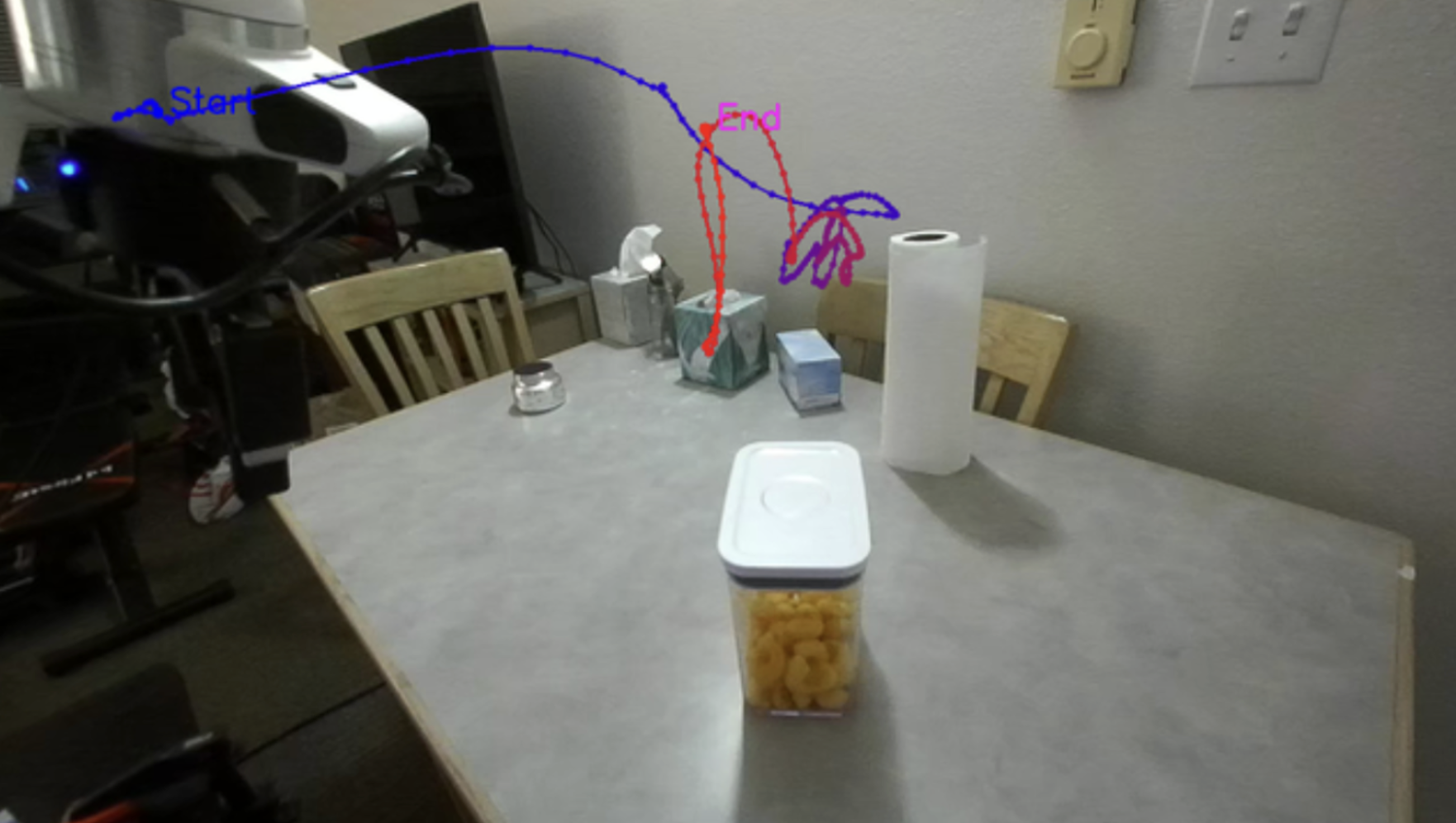}%
{Which language instruction best describes the robot's trajectory shown in the image?\\
\emph{[Move the container to the tray, Push the pen towards the bin, Align the box with the drawer, open the container lid, Lift the cup upward]}}%
{Open the container lid}%
{Answers involving absent objects (pen, cup) were quickly eliminated. The trajectory clearly aligned with the container, making the correct answer straightforward.}%
{Robot trajectory clearly aligned with opening a container lid, excluding irrelevant options involving absent items.}

\vqafigure{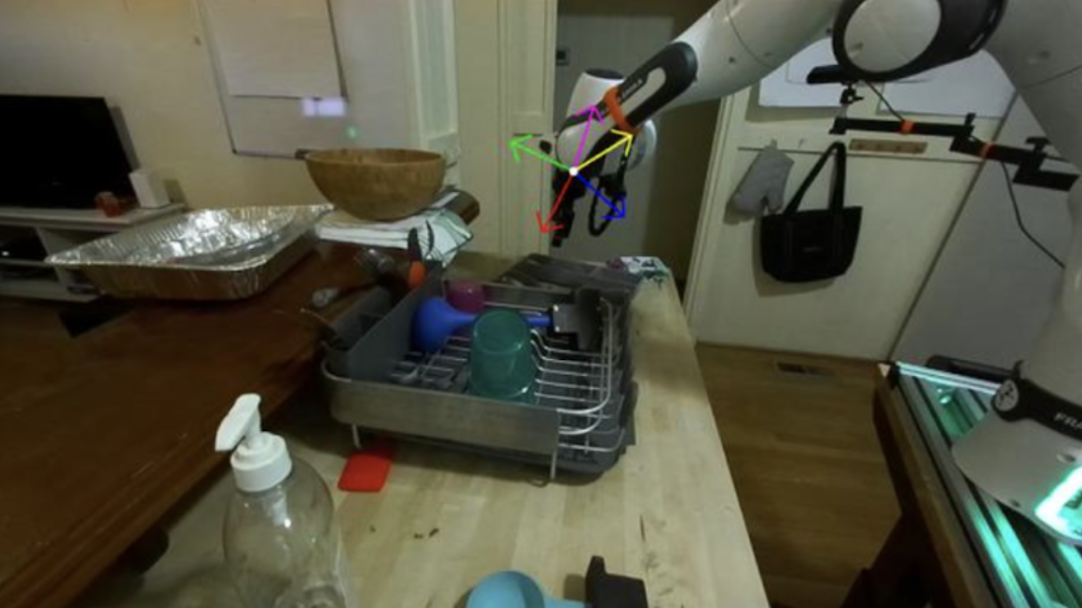}%
{The robot task is to move the spoon. Which colored arrow shows the most likely direction the robot will move next?\\
\emph{[Yellow, Purple, Blue, Green, Red]}}%
{Red}%
{Initially unclear about the spoon's exact position, I carefully inspected to confirm the gripper already grasped the spoon, identifying the red arrow direction correctly.}%
{Discerning the direction of spoon movement based on visual cues, highlighting careful visual analysis.}

\vqafigure{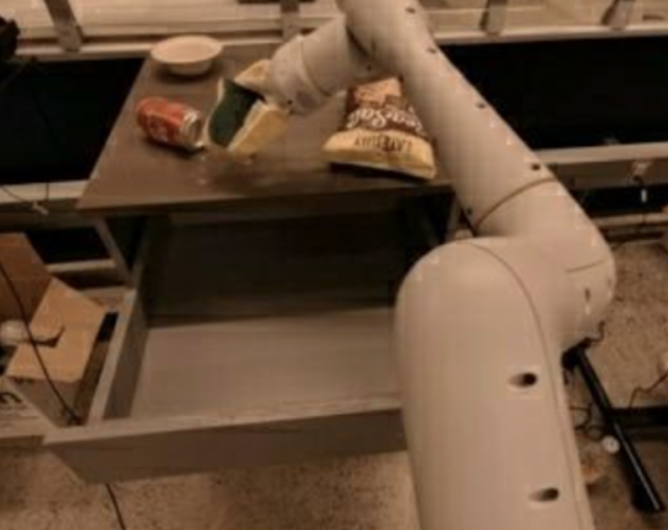}%
{Is the robot's grasp of the sponge stable?\\
\emph{[Yes, No, Cannot be determined, Partially stable]}}%
{No}%
{At first glance, the grip seemed stable, but closer examination revealed the grasp was inadequate on the sponge's edge, confirming instability.}%
{Evaluating the stability of a robotic grasp on a sponge, emphasizing close visual inspection to determine grasp quality.}

\vqafigureduo{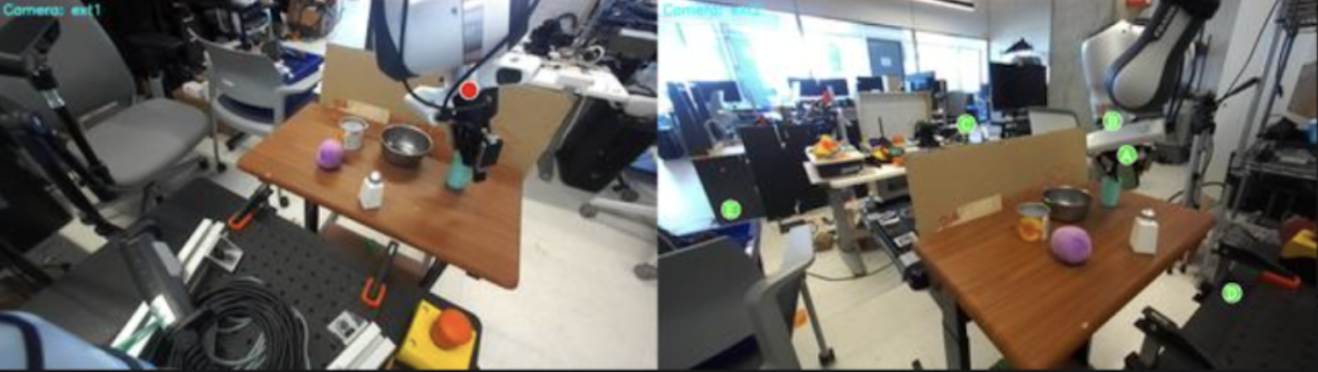}%
{In the left image (ext1 camera), a red dot is marked. Which point in the right image (ext2 camera) corresponds closest to this dot?\\
\emph{[A, B, C, D]}}%
{D}%
{Distinguishing between similarly close points (A and B) required careful analysis. By comparing unique features (such as the wrist camera and the joint’s white part), the correct point became evident.}%
{Identifying corresponding points between two camera views, requiring detailed analysis of visual similarities.}

\vqafigure{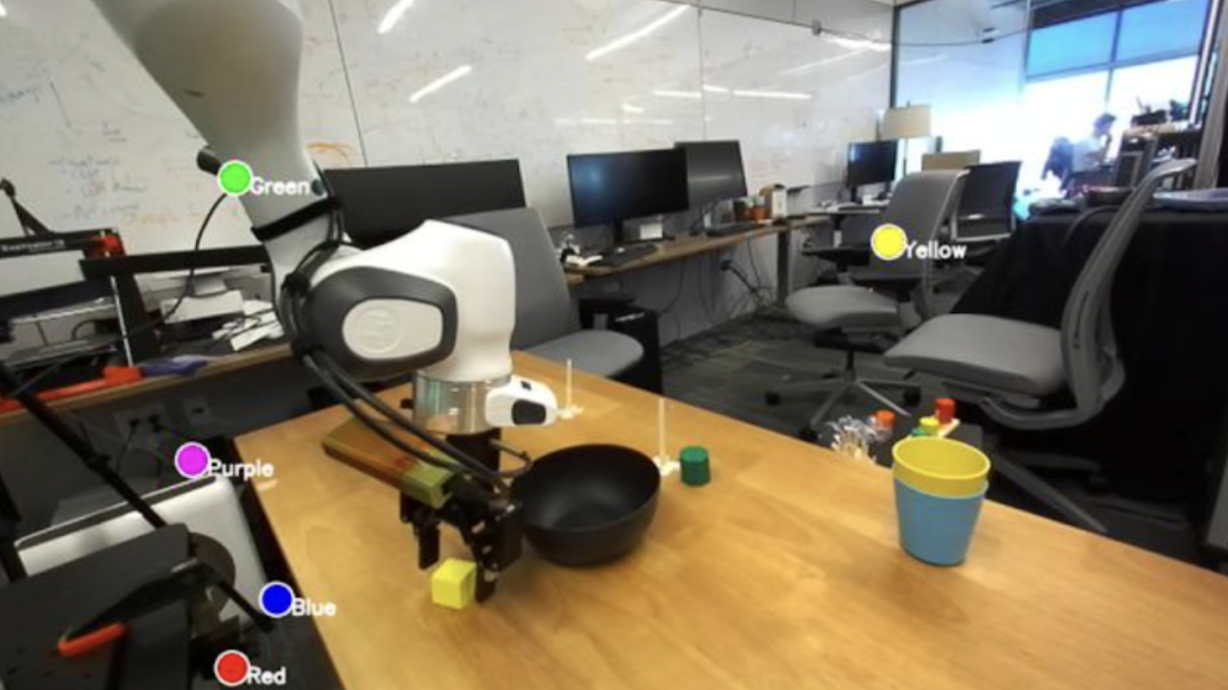}%
{In the ext2 camera image, which colored point is closest to the camera?\\
\emph{[A, B, C, D, E]}}%
{C}%
{Initially, I assumed the blue point was on the table, but a second look revealed it was on the floor, indicating clearly that point C was closest to the camera.}%
{Illustrating depth perception from camera viewpoints and correcting initial visual assumptions upon closer inspection.}

\section{VLM Output Examples}

\begin{figure}[H]
    \centering
    \includegraphics[width=\linewidth]{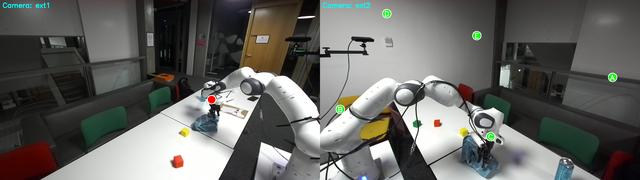}
    \caption{\textbf{Question}: In the left image (ext1 camera), a red dot is marked. Which point is the closest point in the right image (ext2 camera) corresponding to the same 3D location? The correct answer is \textbf{D. point C}.}
    \label{fig:enter-label}
\end{figure}

\begin{tcolorbox}[colback=blue!3!white, colframe=blue!30!black,
  title=Model: \texttt{llava-hf/llava-1.5-7b-hf}, fonttitle=\bfseries, breakable]

\textbf{Prediction:} The closest point in the right image (ext2 camera) corresponding to the same 3D location as the red dot in the left image (ext1 camera) is \textbf{point B}.

\vspace{0.5em}
\textbf{Step-by-step explanation:}
\begin{itemize}
    \item In the left image, the red dot is marked at the top-left corner.
    \item In the right image, point B is also located in the top-left corner.
    \item The red dot's position is consistent across both views.
    \item Thus, point B corresponds to the same 3D location as the red dot.
\end{itemize}

\textbf{Final Answer:} \texttt{B} \xmark
\end{tcolorbox}

\begin{tcolorbox}[colback=blue!3!white, colframe=blue!30!black,
  title=Model: \texttt{Qwen/Qwen2.5-VL-7B-Instruct}, fonttitle=\bfseries, breakable]

\textbf{Prediction:} To determine the corresponding point in the right image, we analyze spatial relationships in both views.

\vspace{0.5em}
\textbf{Step-by-step explanation:}
\begin{itemize}
    \item The red dot in the left image is on the table, slightly left of center and near the bottom.
    \item In the right image, a small red object appears in the same relative position.
    \item Candidate analysis:
    \begin{itemize}
        \item A: Point B — too far from the red object.
        \item B: Point A — located on the wall.
        \item C: Point E — not aligned spatially.
        \item D: Point C — best matches the red dot position.
        \item E: Point D — farther than Point C.
    \end{itemize}
\end{itemize}

\textbf{Final Answer:} \texttt{D} \cmark
\end{tcolorbox}

\begin{tcolorbox}[colback=blue!3!white, colframe=blue!30!black,
  title=Model: \texttt{meta-llama/Llama-3.2-11B-Vision-Instruct}, fonttitle=\bfseries, breakable]

\textbf{Prediction:} Point \textbf{E} in the right image corresponds to the same 3D location as the red dot in the left image.

\vspace{0.5em}
\textbf{Step-by-step explanation:}
\begin{itemize}
    \item Locate the red dot in the left image (ext1 camera) — it's on the table in front of the robot.
    \item In the right image (ext2 camera), observe all labeled points.
    \item Point E is spatially aligned with the red dot’s 3D location.
\end{itemize}

\textbf{Final Answer:} \texttt{E} \xmark
\end{tcolorbox}

\begin{figure}[H]
    \centering
    \includegraphics[width=\linewidth]{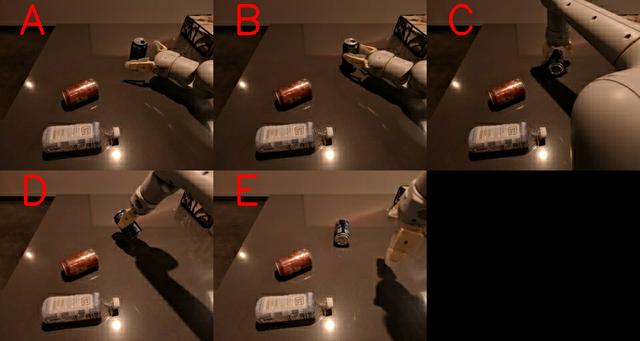}
    \caption{\textbf{Question}: The robot's task is to place pepsi can upright. Which configuration shows the goal state that the robot should achieve? [ "Configuration D", "Configuration E", "Configuration B", "Configuration C", "Configuration A" ]}
    \label{fig:enter-label}
\end{figure}

\begin{tcolorbox}[colback=blue!3!white, colframe=blue!30!black,
  title=Model: \texttt{llava-hf/llava-v1.6-mistral-7b-hf}, fonttitle=\bfseries, breakable]

\textbf{Prediction:}
To determine the configuration that shows the goal state for the robot, we consider:
\begin{itemize}
    \item The task requires placing Pepsi cans upright.
    \item The robot starts with cans on their sides.
    \item We must identify the configuration where the cans are fully upright.
\end{itemize}

\textbf{Analysis of configurations:}
\begin{itemize}
    \item \textbf{Configuration A:} One Pepsi can is upright; two are on their sides.
    \item \textbf{Configuration B:} All cans are on their sides.
    \item \textbf{Configuration C:} One can is upright; two are on their sides.
    \item \textbf{Configuration D:} Two cans upright, two on their sides.
    \item \textbf{Configuration E:} All cans are on their sides.
\end{itemize}

\textbf{Final Answer:} \texttt{A} \xmark

\end{tcolorbox}

\begin{tcolorbox}[colback=blue!3!white, colframe=blue!30!black,
  title=Model: \texttt{Qwen/Qwen2.5-VL-7B-Instruct}, fonttitle=\bfseries, breakable]

\textbf{Prediction:}
To determine which configuration shows the goal state that the robot should achieve, let's analyze the sequence of images step by step:
\begin{itemize}
    \item \textbf{Configuration A:} The Pepsi can is initially lying on its side.
    \item \textbf{Configuration B:} The can has been moved to a vertical position.
    \item \textbf{Configuration C:} The can remains vertical but is shifted slightly to the right.
    \item \textbf{Configuration D:} The can is still transitioning toward an upright position.
    \item \textbf{Configuration E:} The can is upright and appears stable.
\end{itemize}

Given the goal of achieving a final, stable upright position:
\begin{itemize}
    \item B and D are transitional.
    \item C is upright but not final.
    \item E is upright and stable.
\end{itemize}

\textbf{Final Answer:} \texttt{E} \cmark

\end{tcolorbox}

\begin{tcolorbox}[colback=blue!3!white, colframe=blue!30!black,
  title=Model: \texttt{meta-llama/Llama-3.2-11B-Vision-Instruct}, fonttitle=\bfseries, breakable]
\textbf{Prediction:} The task is to place the Pepsi can upright.  
\begin{itemize}
    \item \textbf{Configuration A:} The can is already upright, which the model assumes is the goal state.
    \item \textbf{Configurations B–E:} The model states the can is still lying on its side in each.
\end{itemize}

\textbf{Conclusion:} The model selects Configuration A as the goal state.

\vspace{0.5em}
\textbf{Final Answer:} \texttt{A} \xmark

\end{tcolorbox}


\begin{figure}[H]
    \centering
    \includegraphics[width=\linewidth]{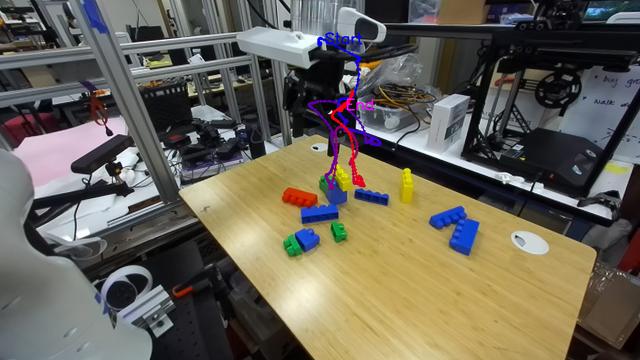}
    \caption{\textbf{Question}: The robot's task is to place pepsi can upright. Which configuration shows the goal state that the robot should achieve? [ "Configuration D", "Configuration E", "Configuration B", "Configuration C", "Configuration A" ]}
    \label{fig:enter-label}
\end{figure}

\begin{tcolorbox}[colback=blue!3!white, colframe=blue!30!black,
  title=Model: \texttt{llava-hf/llava-v1.6-mistral-7b-hf}, fonttitle=\bfseries, breakable]

\textbf{Prediction:}
The model notes uncertainty due to limited visual context and evaluates each option:
\begin{itemize}
    \item A – Rejected: no tower visible in the image.
    \item B – Possible but lacks visual confirmation of gripper state.
    \item C – Rejected: claimed to be irrelevant (note: this is the correct answer).
    \item D – Selected: plausible if the robot is about to pick up a block.
    \item E – Rejected: no tower present.
\end{itemize}

Based on this reasoning, the model selects:
\textbf{Final Answer:} \texttt{D} \xmark

\end{tcolorbox}

\begin{tcolorbox}[colback=blue!3!white, colframe=blue!30!black,
  title=Model: \texttt{Qwen/Qwen2.5-VL-7B-Instruct}, fonttitle=\bfseries, breakable]

\textbf{Prediction:}
The model analyzes the trajectory and visual setup:
\begin{itemize}
    \item The robot's arm is extended toward a block on the table.
    \item The movement suggests an upcoming grasp action.
    \item The gripper is not yet in contact with the object.
\end{itemize}

\textbf{Reasoning:}
\begin{itemize}
    \item A – No rotation or tower is observed.
    \item D – The object is not on the floor.
    \item E – No tray is visible.
    \item C – Incorrectly dismissed, despite covering the multi-step nature of the trajectory.
\end{itemize}

The model concludes the most likely intent is to grab the block:

\textbf{Final Answer:} \texttt{B} \xmark

\end{tcolorbox}

\begin{tcolorbox}[colback=blue!3!white, colframe=blue!30!black,
  title=Model: \texttt{meta-llama/Llama-3.2-11B-Vision-Instruct}, fonttitle=\bfseries, breakable]
\textbf{Prediction:} The robot’s trajectory involves multiple steps:
\begin{itemize}
    \item Rotating the tower clockwise
    \item Grabbing a block with the gripper
    \item Picking up the block from the floor
    \item Dropping the tower into a tray
\end{itemize}

Each option in the list describes only a partial action:
\begin{itemize}
    \item A – Rotation only
    \item B – Grabbing only
    \item D – Picking up only
    \item E – Dropping only
\end{itemize}

Since no single instruction fully describes the trajectory, the model selects:
\textbf{Answer:} \texttt{C. None of the above} \cmark

\end{tcolorbox}
\section{Prompt Design for VQA Inference and Verification}
This section presents the prompt design strategies used for three key VQA inference settings: answer verification, zero-shot inference, and chain-of-thought (CoT) reasoning. The verifier prompt guides a model to extract and isolate the correct multiple-choice answer from a generated explanation, ensuring alignment between reasoning and final answer format. The zero-shot prompt enforces concise behavior by instructing the model to output only the letter corresponding to the correct answer without additional reasoning. In contrast, the CoT prompt encourages step-by-step reasoning before concluding with the final answer, enabling the model to explain its decision-making process. Additionally, Table~\ref{tb:prompt} outlines prototype question types used in Robo2VLM. 

\subsection{Prompt for Verifier}

The verifier prompt is used to post-process model-generated answers that contain free-form text, such as in CoT or long-form reasoning outputs. It instructs the model (or another lightweight parser model) to extract the final answer option—typically a letter (A, B, C, D, or E)—from the full response. This prompt plays a critical role in decoupling reasoning quality from answer accuracy, allowing us to evaluate whether the model reaches a correct conclusion after potentially verbose reasoning. The design includes an illustrative example to make the extraction instruction explicit and reduce hallucination of unexpected formats.






\begin{tcolorbox}[colback=green!3!white, colframe=green!30!black, 
  title=Example: Verifier Prompt, fonttitle=\bfseries, sharp corners=south]
\textbf{Instructions:} Please read the example below and extract the final answer from the model response.  

\textit{Hint:} Your output should be a single letter (e.g., A, B, C, or D) indicating the correct option.

\vspace{1em}
\textbf{Question:} What fraction of the shape is blue?  

\textbf{Choices:} (A) 3/11 \hspace{1em} (B) 8/11 \hspace{1em} (C) 6/11 \hspace{1em} (D) 3/5

\vspace{1em}
\textbf{Model response:} The correct answer is (B) 8/11.

\vspace{1em}
\textbf{Extracted answer:} \texttt{B}
\end{tcolorbox}

\subsection{Prompt for Zero-Shot}

The zero-shot prompt is optimized for direct evaluation of pretrained VLMs without any in-context demonstrations. It instructs the model to select one option from a multiple-choice question using only the corresponding letter. The prompt avoids any reasoning cues or explanations, forcing the model to rely entirely on its pretrained visual and language priors. This prompt setting allows us to assess the model’s default grounding and answer formulation capabilities, free from inductive biases introduced by reasoning scaffolds.

\begin{tcolorbox}[colback=green!3!white, colframe=green!30!black, 
  title=Prompt: Zero-Shot Inference, fonttitle=\bfseries, sharp corners=south]
\textbf{Instructions:} Answer the following multiple-choice question by selecting the correct option letter only.

\textit{Hint:} Do not include any explanation—your response should only contain one of the letters: A, B, C, D, or E.

\{Question\}
\end{tcolorbox}
The \{question\} can be found in Table ~\ref{tb:prompt}.

\subsection{Prompt for Chain-of-Thoughts}

To improve performance on questions that benefit from intermediate reasoning steps (e.g., spatial inference, task planning, or temporal prediction), we adopt a CoT prompt that encourages step-by-step explanation before committing to a final answer. The CoT prompt explicitly requests both reasoning and a conclusive answer in a standard format, helping the model avoid trailing off or omitting a definitive choice. This setting is particularly useful for analyzing the internal decision-making process of large language models in complex manipulation scenarios.

\begin{tcolorbox}[colback=green!3!white, colframe=green!30!black, 
  title=Prompt: Chain-of-Thought Reasoning, fonttitle=\bfseries, sharp corners=south]
\textbf{Instructions:} Answer the following multiple-choice question by reasoning step by step. Show your work for each step before concluding.

\textit{Hint:} After completing your reasoning, output only the final answer option letter (A, B, C, D, or E) at the end.

\{Question\}
\end{tcolorbox}
The \{question\} can be found in Table ~\ref{tb:prompt}.

\begin{table}[ht]
\centering
\caption{Question Prompt Templates for VQA Functions}
\begin{tabular}{p{0.3\textwidth} p{0.65\textwidth}}
\toprule
\textbf{VQA Function} & \textbf{Question Prompt Prototype} \\
\midrule
\texttt{robot\_gripper\_open} & Is the robot's gripper open? \\
\addlinespace
\texttt{object\_reachable} & Is there any obstacle blocking the robot from reaching \{object\}? \\
\addlinespace
\texttt{relative\_direction} & In the image from \{camera\} at step \{step\},
which direction is the \{object\} relative to the robot's end effector? \\
\addlinespace
\texttt{relative\_depth} & In the image from \{camera\},
which colored point is closest/farthest from the camera? \\
\addlinespace
\texttt{view\_correspondence} & In the left image (\{camera1\}), a red dot is marked.
Which point in the right image (\{camera2\}) corresponds to the same location? \\
\addlinespace
\texttt{task\_success\_state} & The robot is to \{instruction\}.
Has the robot successfully completed the task? \\
\addlinespace
\texttt{is\_stable\_grasp} & Is the robot's grasp of the \{object\} stable? \\
\addlinespace
\texttt{goal\_configuration} & The robot's task is to \{instruction\}.
Which configuration shows the goal state? \\
\addlinespace
\texttt{action\_understanding} & The robot is tasked to \{instruction\}.
Which phase of the grasp action is shown? \\
\addlinespace
\texttt{next\_action} & After \{current phase\},
what will be the robot's NEXT action phase? \\
\addlinespace
\texttt{trajectory\_understanding} & Which language instruction best describes
the robot's trajectory shown in the image? \\
\addlinespace
\texttt{action\_direction} & Which colored arrow correctly shows the direction
the robot will move next? \\
\addlinespace
\texttt{temporal\_sequence} & What is the correct sequence of action phases
shown in the images? \\
\bottomrule
\end{tabular}
\label{tb:prompt}
\end{table}

\newpage
\section{Fine-Tuning and Evaluation Details}
\subsection{Fine-Tuning Details}

\paragraph{Model Configuration} The model utilized for vision-language tasks is based on \texttt{meta-llama/Llama-3.2-11B-Vision}, configured for optimal performance. Key settings include gradient checkpointing with the "unsloth" method, a LoRA (Low-Rank Adaptation) rank of 128, an alpha parameter of 256, and no dropout for LoRA modules. Model fine-tuning is selectively enabled for language layers, attention modules, and MLP modules while keeping vision layers fixed. The maximum sequence length is set to 2048 tokens to accommodate complex vision-language interactions.

\paragraph{Training Setup} Training utilizes the dataset \texttt{keplerccc/ManipulationVQA-60k} with a dedicated train split and a validation ratio of 5\%. Batch size is carefully controlled at 4 samples per device, enhanced by gradient accumulation over 4 steps. The training process involves linear scheduling of the learning rate, starting at , and includes a weight decay of 0.01. The training is configured to run for one epoch with frequent checkpoints every 1000 steps, evaluation intervals at 5000 steps, and logging every 10 steps.

\paragraph{Evaluation Protocol} Evaluation is conducted using a maximum of 10,000 test samples, with explicit configuration for generating visualizations and fallback strategies in case of missing test splits. Generation settings include sampling with a temperature of 0.7 and allowance of up to 50 new tokens per generation. The evaluation setup includes assessing both base and fine-tuned model versions, each clearly delineated within the configuration.

\paragraph{Distributed Training and Precision} The system leverages distributed training techniques, exploiting high-performance computational resources for scalable training. It utilizes Brain Floating Point (BF16) precision to balance computational efficiency and numerical stability, eschewing FP16 for better performance stability.

\subsection{Evaluation Details}
\paragraph{Experimental Setup} We conducted evaluations using a vision-language model (VLM) pipeline configured specifically for Visual Question Answering (VQA) tasks. The evaluation utilizes the Hugging Face dataset named \texttt{keplerccc/ManipulationVQA}, specifically the \texttt{test} split, enabling standardized comparisons. To maintain computational efficiency and manage GPU resources effectively, the evaluation employs adaptive batch processing strategies.

\paragraph{Model Configuration} The evaluation primarily considers two large-scale multimodal models: \texttt{llava-hf/llava-v1.6-34b-hf} and \texttt{llava-hf/llava-next-72b-hf}. These models leverage tensor parallelism set to 4, harnessing the full computational power of four A100 GPUs to optimize throughput. The models were initialized with a GPU memory utilization parameter set to 0.9, ensuring efficient memory usage without exceeding GPU capacity.

\paragraph{Prompt and Response Extraction} Each evaluation prompt explicitly instructs the models to select from multiple-choice answers (options A, B, C, D, E). Responses are subsequently processed using a secondary extraction model (\texttt{meta-llama/Llama-3.2-3B-Instruct}), designed to deterministically extract the selected letter-answer from the models' verbose outputs. This extraction leverages zero-temperature sampling to guarantee reproducibility and consistency across evaluations.

\paragraph{Dataset and Evaluation Metrics} The dataset comprises a randomly shuffled subset of test questions, limited by a configurable maximum sample parameter. Accuracy metrics are computed overall and further broken down by tags to provide granular insights into model performance across different question categories. Detailed timing information for responses is recorded to assess computational efficiency, reporting average response times alongside accuracy metrics.

\section{Human Expert Instruction and Feedback} \label{sec:human-expert}

To improve the quality and answerability of automatically generated questions, we ask a human expert to improve the data generation process. We provided an initial set of 200 question-image pairs generated by the Robo2VLM pipeline to a human expert for review. The expert was instructed to identify unanswerable or ambiguous cases and annotate the reasons, which were then used to iteratively refine the prompt and generation pipeline. The human expert takes two hours to complete the evaluation. We then follow the revised questions to generate the whole dataset. 

\subsection{Evaluation Protocol}

The human expert was asked to assess whether each question could be reliably answered based solely on the visual input and accompanying instruction. For cases deemed unanswerable, the expert selected from predefined failure modes including: (1) insufficient or unclear visual context, (2) ambiguous or underspecified language in the prompt, and (3) other task-specific issues. This structured feedback guided the refinement of question templates, robot state annotations, and visual preprocessing steps.

\subsection{Feedback-Driven Refinement of Auto-Curation}

Table~\ref{tab:problem_solutions} summarizes the key issues uncovered through human evaluation and the corresponding solutions incorporated into the Robo2VLM pipeline. These challenges fall into four main categories: (i) \textit{Context and Task Definition}, addressing missing goal descriptions and task phase awareness; (ii) \textit{Visual Information and Camera Limitations}, such as limited visibility or poor resolution, which were mitigated through multi-view integration and filtering heuristics; (iii) \textit{Question Formulation and Consistency}, where we standardized linguistic structures, unified success criteria, and added consistency validation checks; and (iv) \textit{Category-Specific Issues}, including configuration reasoning, spatial alignment, and directional prediction, which were resolved using goal-aware and multi-perspective analysis. Together, these improvements enhance the reliability, interpretability, and generalization of vision-language evaluations in robotic settings.

\begin{table*}[t]
\centering
\caption{Problems Identified by Human Experts and Corresponding Solutions Implemented in Robo2VLM Pipeline}
\footnotesize
\label{tab:problem_solutions}
\begin{tabular}{p{0.47\textwidth}|p{0.47\textwidth}}
\toprule
\textbf{Problem Category} & \textbf{Implemented Solution} \\

\rowcolor{gray!15} \multicolumn{2}{l}{\textbf{Context and Task Definition}} \\

Image understanding issue without task context & Enhanced question prompt with task context \\
\addlinespace
Lack of goal specificity & Enhanced question prompt with goal descriptions \\
\addlinespace
Assumed implicit knowledge of robotic tasks & Added description of the robot's current phase \\

\rowcolor{gray!15} \multicolumn{2}{l}{\textbf{Visual Information and Camera Limitations}} \\

Limited wrist camera view and object visibility & Integrated multi-view images \\
\addlinespace
Invisible gripper state from certain angles & Added gripper state verification and filtering \\
\addlinespace
Insufficient image resolution for detailed object identification & Filtered out images with resolution lower than 100$\times$100 pixels \\

\rowcolor{gray!15} \multicolumn{2}{l}{\textbf{Question Formulation and Consistency}} \\

Ambiguous or complex question phrasing & Standardized linguistic templates \\
\addlinespace
Inconsistent task completion criteria & Unified success state definitions \\
\addlinespace
Redundant or confusing phrasing & Applied phrase filtering and clarity scoring \\
\addlinespace
Conflicting answers across questions for same image & Added consistency validation checks \\

\rowcolor{gray!15} \multicolumn{2}{l}{\textbf{Category-Specific Issues}} \\

Multiple viewpoints needed for configuration selection & Added multi-angle verification to configuration questions \\
\addlinespace
Spatial reasoning depends on object boundaries and color & Improved spatial questions with object detection and color validation \\
\addlinespace
Direction prediction depends on task goal & Integrated goal-aware motion prediction \\
\bottomrule
\end{tabular}
\end{table*}


\section{Key dataset statistics}
We analyzed a total of 60,000 samples in the dataset. On average, questions are 108.69 characters long, with a median length of 113 characters. The shortest question contains 28 characters, while the longest reaches 378 characters. Each question includes an average of 4.65 answer choices, with most having either 4 or 5 options. The typical choice is 14.22 characters long on average, though lengths vary widely—from as short as 1 character to as long as 271 characters. The combined length of all choices per question averages 66.09 characters, with a median of 44 characters and a range from 5 to 687 characters.

In terms of correct answer distribution, the dataset is relatively balanced among options A to D: 22.03\% of correct answers are 'D', 21.86\% are 'B', 21.74\% are 'C', and 21.53\% are 'A'. Option 'E' appears less frequently, making up 12.84\% of correct responses.

Regarding image data, the average image width is 520.66 pixels, with a median of 640 pixels, while heights average 292.99 pixels, with a median of 256 pixels. Image widths range from 84 to 640 pixels, and heights from 84 to 480 pixels. The most common image resolutions are 640x360 (39.61\%), 320x256 (21.14\%), 640x240 (8.46\%), 640x180 (5.81\%), and 448x224 (4.14\%). Across the dataset, there are 19 unique image resolutions.

\begin{table}[h]
\centering
\footnotesize
\caption{Dataset Statistics Summary for 60,000 Samples}
\begin{tabular}{llr}
\toprule
\textbf{Category} & \textbf{Metric} & \textbf{Value} \\
\midrule

\multirow{4}{*}{Questions} 
  & Avg. length (chars) & 108.69 \\
  & Median length (chars) & 113.00 \\
  & Min length (chars) & 28 \\
  & Max length (chars) & 378 \\

\midrule
\multirow{7}{*}{Choices} 
  & Avg. \# choices per question & 4.65 \\
  & Median \# choices & 5.00 \\
  & Min \# choices & 4 \\
  & Max \# choices & 5 \\
  & Avg. length of a choice (chars) & 14.22 \\
  & Median length of a choice (chars) & 6.00 \\
  & Min/Max choice length (chars) & 1 / 271 \\
\midrule
\multirow{3}{*}{Choices (total per question)} 
  & Avg. total length (chars) & 66.09 \\
  & Median total length (chars) & 44.00 \\
  & Min/Max total length (chars) & 5 / 687 \\

\midrule
\multirow{5}{*}{Answer Distribution} 
  & A & 12,918 (21.53\%) \\
  & B & 13,115 (21.86\%) \\
  & C & 13,046 (21.74\%) \\
  & D & 13,216 (22.03\%) \\
  & E & 7,705 (12.84\%) \\

\midrule
\multirow{4}{*}{Image Width (px)} 
  & Avg. & 520.66 \\
  & Median & 640.00 \\
  & Min/Max & 84 / 640 \\

\midrule
\multirow{4}{*}{Image Height (px)} 
  & Avg. & 292.99 \\
  & Median & 256.00 \\
  & Min/Max & 84 / 480 \\

\midrule
\multirow{6}{*}{Top-5 Resolutions} 
  & 640x360 & 23,768 (39.61\%) \\
  & 320x256 & 12,683 (21.14\%) \\
  & 640x240 & 5,075 (8.46\%) \\
  & 640x180 & 3,484 (5.81\%) \\
  & 448x224 & 2,482 (4.14\%) \\
  & Unique resolutions & 19 \\
\bottomrule
\end{tabular}
\end{table}

\section{Distractor Choice Design}
This section outlines the design and evaluation of distractor choices in our VQA dataset, which play a critical role in determining question difficulty and diagnostic value. We begin by examining the impact of introducing a ``None of the Above'' (NAB\%) option, which systematically increases task ambiguity and reduces model performance across the board (Fig.~\ref{fig:noab}). We then detail the principles and heuristics used to generate diverse and context-aware distractors for different question types. These include binary negations, categorical sampling, spatial reasoning perturbations, and content-aware language distractors. Special emphasis is placed on generating plausible incorrect choices that reflect partial knowledge, ambiguity, or visually confusable elements. Finally, we describe how randomized shuffling and probabilistic replacement with NAB options further strengthen the challenge by discouraging rote pattern matching. Together, these strategies enhance the dataset’s ability to probe fine-grained reasoning, visual grounding, and robustness to uncertainty in large vision-language models.

\subsection{None of the Above Proportion}
\begin{figure}[H]
    \centering
    \includegraphics[width=0.6\columnwidth]{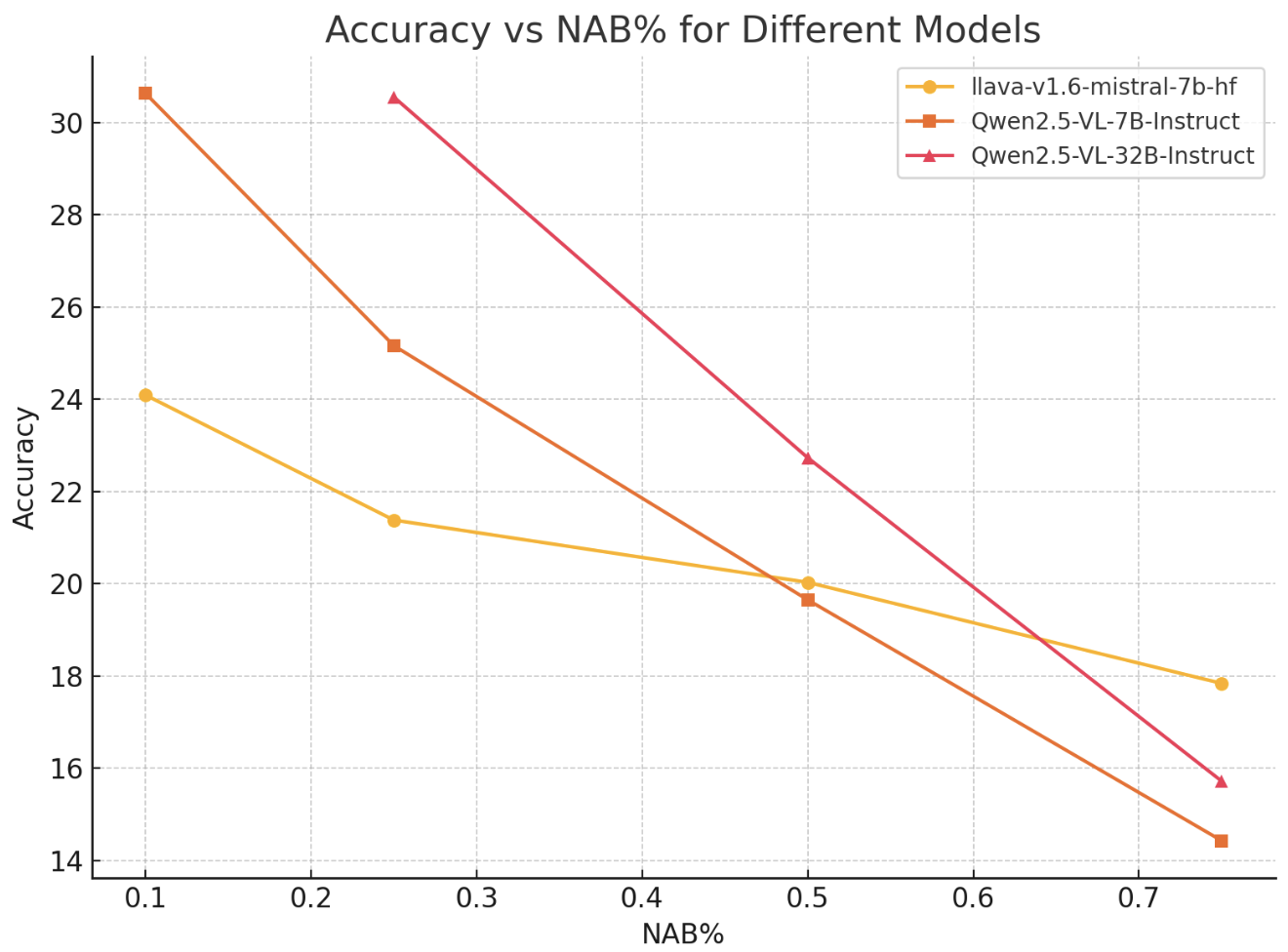}
    \caption{Accuracy vs. ‘None of the Above’ Selection Ratio (NAB\%) for Three Vision-Language Models }
    \label{fig:noab}
\end{figure}

This section shows experiment of adding ‘None of the Above’ selection Ratio (NAB\%) choice increase the difficulty of the dataset and model accuracy decrease for all the models. We show the result in the line plot in Fig.~\ref{fig:noab}.

The plot reveals that all three models experience a decline in accuracy as NAB\% increases, indicating reduced confidence or higher prediction difficulty when a greater proportion of questions are considered potentially unanswerable. Qwen2.5-VL-32B-Instruct consistently outperforms the other two models when data is available, achieving the highest accuracy of 30.55\% at NAB\% = 0.25. Interestingly, the 7B Qwen2.5-VL variant initially performs well (30.63\% at NAB\% = 0.1) but degrades more sharply than the 32B version. The llava-v1.6-mistral-7b-hf model maintains the lowest accuracy across all NAB\% levels, suggesting it is less robust under ambiguity. These trends highlight the importance of model scale and training data in handling tasks with varying uncertainty.

\subsection{Distractors}
The design of distractor choices is crucial for creating challenging and meaningful Visual Question Answering (VQA) instances. The provided Python codebase employs several strategies to generate plausible yet incorrect options, aiming to test nuanced understanding rather than simple pattern recognition.

\paragraph{Binary and Generic Distractors}
For questions anticipating a binary response (e.g., Yes/No), the primary distractor is often the direct negation of the correct answer. This is evident in functions like \texttt{vqa\_robot\_gripper\_open} and \texttt{vqa\_object\_reachable}. These are typically supplemented by generic distractors such as ``Cannot be determined'' or context-specific but still general alternatives like ``Partially open'' or ``Partially reachable''. The \texttt{\_validate} method ensures that binary questions have exactly four choices, accommodating these patterns.

\paragraph{Categorical and Permutation-Based Distractors}
Many VQA generation functions define a set of possible categories and select distractors from those not matching the correct answer.
\textbf{Relative Directions:} In \texttt{vqa\_relative\_direction}, a comprehensive list of possible spatial relations (e.g., ``Upper Left'', ``Lower Forward'') is generated. After identifying the correct direction, incorrect choices are drawn from this list, with a preference for those sharing some component (e.g., the same vertical component) with the correct answer to increase plausibility.
\textbf{Action Phases:} For \texttt{vqa\_action\_understanding} and \texttt{vqa\_next\_action}, distractors are chosen from a defined set of robot action phase descriptions (e.g., ``Approaching the object with open gripper'', ``Firmly grasping the object''). The incorrect choices are the descriptions of other valid phases.
\textbf{Temporal Sequences:} \texttt{vqa\_temporal\_sequence} generates distractors by creating incorrect orderings (permutations) of the actual sequence of events or phases if the question is about the sequence itself.
\textbf{Color/Label-Based Choices:} In \texttt{vqa\_relative\_depth} and \texttt{generate\_action\_direction\_selection\_vqa}, distinct colors (e.g., ``Red'', ``Green'', ``Blue'') are assigned to different points or arrows in the image. The choices are then these color names, with one corresponding to the correct visual marker. Similarly, \texttt{vqa\_multi\_view\_correspondence} uses letter labels (``A'', ``B'', ``C'', ``D'', ``E'') for choices corresponding to marked points.

\paragraph{Spatially Derived Distractors}
For tasks involving spatial reasoning, distractors are often generated to be distinct in the image space. In \texttt{vqa\_multi\_view\_correspondence}, distractor points are generated in different quadrants of the image from the correct corresponding point, ensuring a minimum pixel distance from each other and the correct point. \texttt{generate\_action\_direction\_selection\_vqa} creates incorrect directional arrows by ensuring their angles are meaningfully different from the correct action direction, with a minimum angular separation.

\paragraph{Content-Based Distractors from External Knowledge}
The \texttt{vqa\_trajectory\_understanding} function generates distractor language instructions by using templates (e.g., ``Pick up the \{\} from the \{\}'') and filling them with common objects and locations, which may or may not be present in the current scene, thus testing a deeper understanding of the visualized trajectory against plausible alternative tasks.

\paragraph{Strategic Shuffling and ``None of the above''}
The \texttt{\_shuffle\_choices} method is systematically called after initial VQA construction. This method randomizes the order of the correct answer and the initially formulated incorrect choices. Furthermore, for non-binary questions (typically those with five choices), there is a 20\% chance to replace the actual correct answer with ``None of the above'', and the original correct answer text is then discarded for that instance, making ``None of the above'' the correct choice. This adds another layer of complexity, requiring the system to not only identify the correct option but also to recognize when none of the substantive options are correct.

The combination of these strategies ensures a diverse set of distractors, tailored to the specific type of question being posed and the visual information presented.
\end{document}